\documentclass{article}

\usepackage{microtype}
\usepackage{graphicx}
\usepackage{amsmath}
\usepackage[dvipsnames]{xcolor}
\usepackage{subcaption}
\usepackage{booktabs} 
\usepackage{multirow}
\usepackage[table]{xcolor}

\usepackage{hyperref}

\usepackage[preprint]{icml2026}

\usepackage{amsmath}
\usepackage{amssymb}
\usepackage{mathtools}
\usepackage{amsthm}
\usepackage{bm}

\usepackage[capitalize,noabbrev]{cleveref}

\theoremstyle{plain}

\theoremstyle{definition}

\theoremstyle{remark}

\usepackage[textsize=tiny]{todonotes}

\icmltitlerunning{Submission and Formatting Instructions for ICML 2026}

\begin{document}

\twocolumn[
  \icmltitle{SAKED: Mitigating Hallucination in Large Vision-Language Models \\ via Stability-Aware Knowledge Enhanced Decoding}

  \icmlsetsymbol{equal}{*}

  \begin{icmlauthorlist}
    \icmlauthor{Zhaoxu Li}{equal,EEE,IJP}
    \icmlauthor{Chenqi Kong}{equal,EEE}
    \icmlauthor{Peijun Bao}{EEE}
    \icmlauthor{Song Xia}{EEE}
    \icmlauthor{Yi Tu}{PH}
    \icmlauthor{Yi Yu}{EEE}

    \icmlauthor{Xinghao Jiang}{SJTU}
    \icmlauthor{Xudong Jiang}{EEE}
  \end{icmlauthorlist}

  \icmlaffiliation{EEE}{ROSE Lab, School of Electrical and Electronic Engineering, Nanyang Technological University, Singapore}
  \icmlaffiliation{IJP}{ROSE Lab, Interdisciplinary Graduate Programme, Nanyang Technological University, Singapore}
  \icmlaffiliation{PH}{School of Physical and Mathematical Sciences, Nanyang Technological University, Singapore}
  \icmlaffiliation{SJTU}{Shanghai Jiao Tong University, China}
  
  \icmlcorrespondingauthor{}{}

  \icmlkeywords{Machine Learning, ICML}
  \vskip 0.3in
]

\printAffiliationsAndNotice{*Equal contribution.}

\begin{abstract}
Hallucinations in Large Vision-Language Models (LVLMs) pose significant security and reliability risks in real-world applications. Inspired by the observation that humans are more error-prone when uncertain or hesitant, we investigate how instability in a model’s internal knowledge contributes to LVLM hallucinations. We conduct extensive empirical analyses from three perspectives, namely attention heads, model layers, and decoding tokens, and identify three key hallucination patterns: (i) visual activation drift across attention heads, (ii) pronounced knowledge fluctuations across layers, and (iii) visual focus distraction between neighboring output tokens. Building on these findings, we propose \textbf{S}tability-\textbf{A}ware \textbf{K}nowledge-\textbf{E}nhanced \textbf{D}ecoding (\textbf{SAKED}), which introduces a layer-wise \textbf{K}nowledge \textbf{S}tability \textbf{S}core (\textbf{KSS}) to quantify knowledge stability throughout the model. By contrasting the most stability-aware and stability-agnostic layers, SAKED suppresses decoding noise and dynamically leverages the most reliable internal knowledge for faithful token generation. Moreover, SAKED is training-free and can be seamlessly integrated into different architectures. Extensive experiments demonstrate that SAKED achieves state-of-the-art performance for hallucination mitigation on various models, tasks, and benchmarks.
\end{abstract}

\section{Introduction}
In recent years, Large Vision–Language Models (LVLMs) have made substantial progress in cross-modal understanding and generation. By connecting a pretrained language model with a vision encoder and fine-tuning on instruction-based datasets, LVLMs achieve strong performance across a wide range of multimodal tasks \cite{openai2023gpt4, chen2023shikra, yin2024survey, zhao2024harmonizing}. Nevertheless, LVLMs remain prone to hallucinations \cite{rohrbach2018object, zhou2023analyzing, li2023evaluating}, where models generate content that is not grounded in the input signals but instead relies on their internal knowledge \cite{heiman2025factchexcker}. Such failures raise pressing safety and security concerns in high-stakes applications, such as autonomous driving, medical diagnosis, and financial analysis, etc \cite{li2025saver, zhang2023language, xia2026provable}.

Prior work has investigated the causes of LVLM hallucinations, including language priors \cite{wang2024mllm, li2025saver}, dataset bias \cite{li2023evaluating, zhou2023analyzing}, error accumulation \cite{huang2024opera, zhou2023analyzing}, and suboptimal training policy \cite{kalai2025language, yang2025mitigating}. Accordingly, a variety of mitigation strategies have been proposed, such as contrastive decoding \cite{leng2024mitigating, favero2024multi, chuang2023dola}, fine-tuning with improved learning policies \cite{jiang2024hallucination, yang2025mitigating, sarkar2024mitigating}, early intervention \cite{wang2024mllm, li2025saver, wang2025shift}, information steering \cite{li2025hidden, park2025steer}, and representation editing \cite{liu2024paying, kang2025see, gong2024damro, kong2025moe}. 
Despite this progress, most existing methods do not explicitly model how internal knowledge fluctuations give rise to hallucinations. Without characterizing when and where internal knowledge becomes unreliable, mitigation methods can be brittle and may fail under distribution shifts \cite{heiman2025factchexcker, lv2024coarse}. In contrast, a stability-aware perspective enables the model to dynamically leverage more reliable internal knowledge, thereby achieving more faithful decoding.

\begin{figure}[ht]
  \begin{center}
  \centerline{\includegraphics[width=\columnwidth]{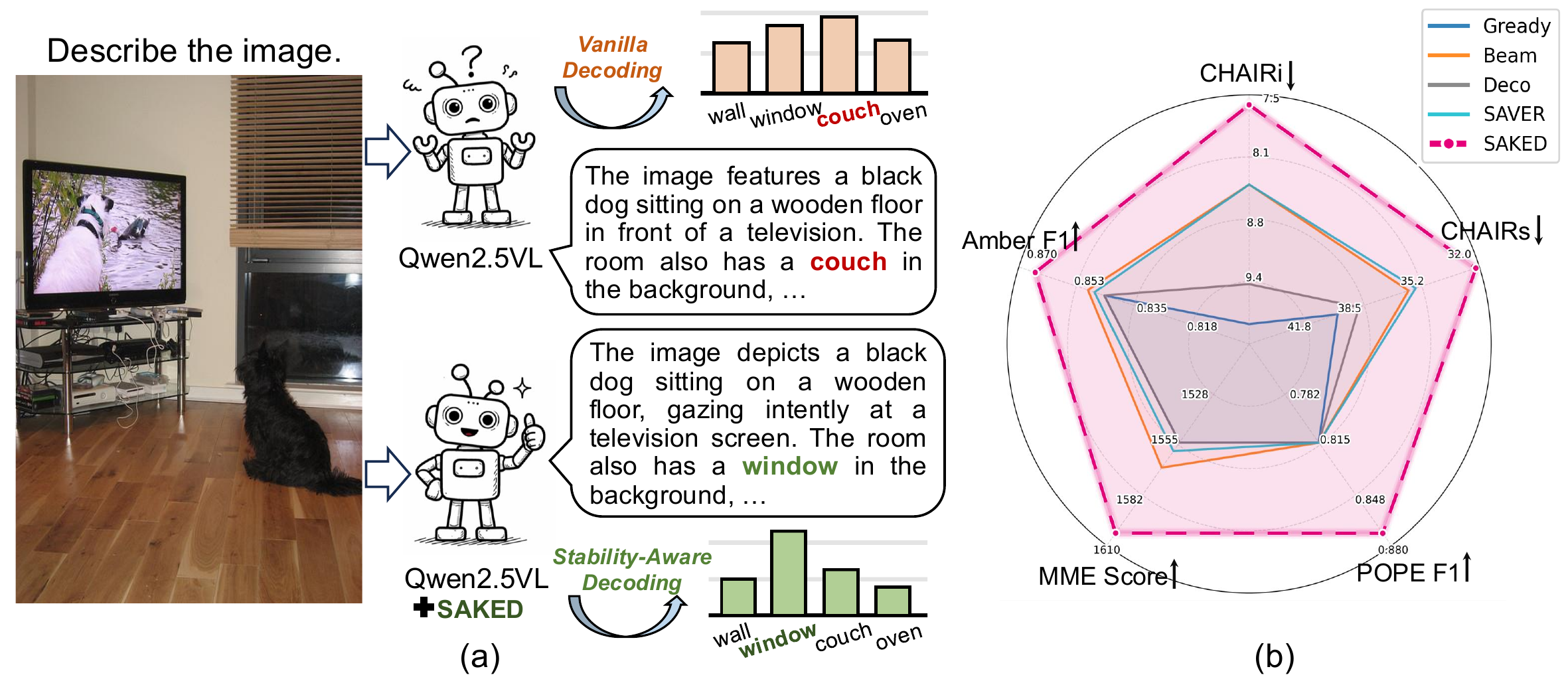}}
    \caption{(a) SAKED mines internal stability-aware knowledge to enhance the decoding process. (b) SAKED consistently achieves outstanding hallucination performance on CHAIR, POPE, MME, and AMBER.
}
    \label{Teaser}
  \end{center}
  \vspace{-6mm}
\end{figure}
Inspired by the observation that humans are more error prone when uncertain or hesitant, we investigate hallucinations in LVLMs through the lens of internal knowledge stability. A key challenge is to explicitly characterize and quantify knowledge fluctuation during generation. To this end, we conduct extensive empirical analyses of knowledge stability from three perspectives: attention heads, model layers, and decoding tokens. These analyses reveal three stability patterns associated with hallucinations. \textbf{Pattern \#1}: Visual activation drifts across attention heads can induce hallucinations; in contrast, heads with consistent visual activations exhibit stronger visual grounding and yield more faithful decoding. \textbf{Pattern \#2}: Large knowledge fluctuations across layers correlate with higher hallucination risk, whereas stable layers preserve more reliable information during token generation. \textbf{Pattern \#3}: Substantial distraction in visual focus between neighboring output tokens also increases likelihood of hallucinations.


We propose {K}nowledge Stability {S}core ({KSS}), which comprises (i) Cross-Head Stability Score ({CHSS}) to measure visual activation drift across attention heads, (ii) Cross-Layer Stability Score ({CLSS}) to quantify knowledge fluctuations across layers, and (iii) Cross-Token Stability Score ({CTSS}) to assess visual focus distraction between adjacent decoding steps. Building on KSS, we further introduce stability-aware contrastive decoding that contrasts logits from the most stability-aware layer and the most stability-agnostic layer to suppress decoding noise and improve  faithfulness. As shown in Figure~\ref{Teaser}, SAKED consistently outperforms prior methods by a clear margin across multiple tasks. 
In summary, our main contributions are as follows:
\vspace{-4mm}
\begin{itemize}
\item We conduct extensive empirical analyses to examine when and where hallucinations arise by tracking model knowledge evolution from three perspectives: attention heads, model layers, and generated tokens. We find that unstable internal model knowledge is associated with a higher risk of hallucination.
\vspace{-1mm}
\item We propose a novel \textbf{S}tability-\textbf{A}ware \textbf{K}nowledge \textbf{E}nhanced \textbf{D}ecoding (\textbf{SAKED}) method that quantifies knowledge stability for each layer. We further improve decoding by contrasting the most stability-aware layer with the most stability-agnostic layer, thereby reducing hallucinations.
SAKED is plug-and-play and can be seamlessly integrated into various LVLMs.
\begin{figure}[ht]
  \begin{center}
  \centerline{\includegraphics[width=\columnwidth]{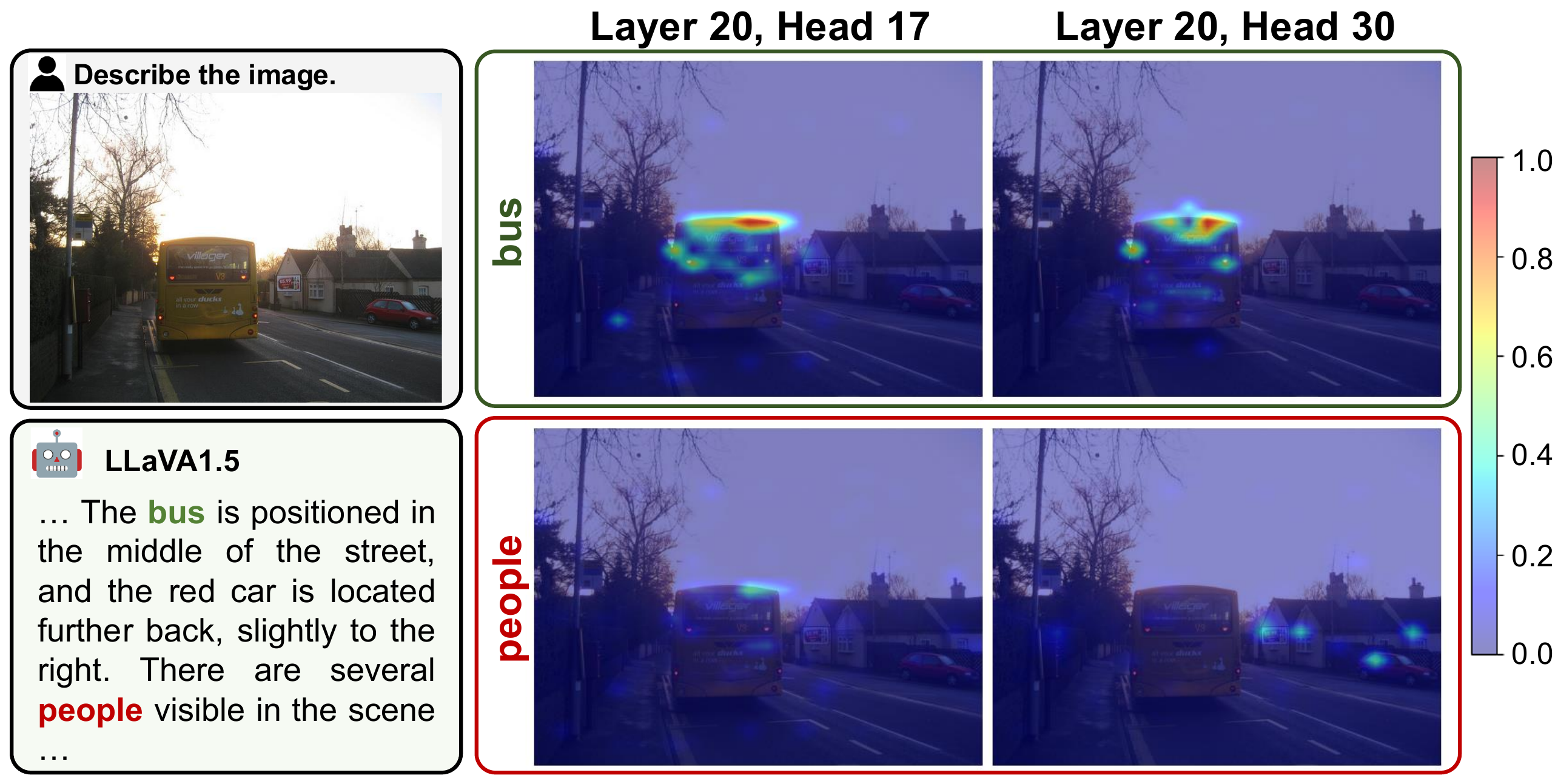}}
    \caption{The left panel shows the input image and the output caption. The right panel visualizes the attention maps of a grounded token (``\textcolor{OliveGreen}{bus}'') and a hallucinated token (``\textcolor{red!70!black}{people}'') across different attention heads. 
    }
    \label{Attention}
  \end{center}
  \vspace{-8mm}
\end{figure}
\vspace{-1mm}
\item Extensive experiments demonstrate that SAKED achieves state-of-the-art hallucination mitigation performance across a wide range of models, datasets, and tasks.
\end{itemize}
\vspace{-3mm}
\section{Related Work}
\vspace{-2mm}
\label{sec:RelatedWork}
\subsection{Large Vision-Language Models (LVLMs)}
Recent advances in LVLMs have yielded strong performance across a wide range of vision–language tasks. Most LVLMs consist of three components: a vision encoder that extracts visual features from the input image, a vision–language connector that aligns these features with the language representation space, and a pretrained LLM that consumes the resulting multimodal embeddings to generate responses. To reduce the modality gap between visual features and textual representations, prior work has explored diverse alignment strategies, including lightweight linear projection layers that map visual features into the LLM token space (e.g., LLaVA~\cite{liu2023visual}, MiniGPT-4~\cite{zhu2023minigpt}), Q-former modules that learn query embeddings for multimodal grounding (e.g., BLIP-2~\cite{li2023blip2}, InstructBLIP~\cite{instructblip}), and cross-attention architectures that enable iterative interaction between visual tokens and text (e.g., Flamingo~\cite{alayrac2022flamingo}, OpenFlamingo~\cite{awadalla2023openflamingo}). More recent systems, such as GPT-5~\cite{openai_introducing_gpt5_2025} and Gemini-3~\cite{gemini3}, further demonstrate enhanced capabilities on vision tasks requiring complex reasoning. Despite this progress, LVLMs remain susceptible to severe hallucinations, raising pressing safety and security concerns, especially in critical application scenarios (e.g., medical diagnosis).

\vspace{-2mm}
\subsection{Hallucination in LVLMs} 
\vspace{-2mm}
Hallucination in LVLMs typically refers to outputs that are not grounded in the input signals and instead rely excessively on the model’s internal knowledge \cite{heiman2025factchexcker}. Recent tuning-based approaches usually improve their learning objectives or training policies \cite{jiang2024hallucination, yang2025mitigating, sarkar2024mitigating, park2025second, xu2024reducing, hesystematic} on high-quality curated datasets \cite{wang2024mitigating, zhang2024reflective, jing2024fgaif}. In contrast, training-free methods are generally more computationally efficient. For example, contrastive decoding \cite{leng2024mitigating, favero2024multi, chuang2023dola, zhualleviating} compares output logits across paired inputs (e.g., different image conditions) to suppress ungrounded generations. Early-intervention approaches \cite{wang2024mllm, li2025saver, wang2025shift, chen2024halc, wu2024evaluating, fang2025grounding, wang2025image, fang2025enhancing} dynamically refine decoding by leveraging more reliable signals from earlier layers. In addition, representation-editing methods \cite{liu2024paying, kang2025see, gong2024damro, zhao2024mitigating, zou2024look, chen2024context, tan2025wisdom} reallocate attention to encourage the model to focus on relevant visual regions. Despite substantial progress, the role of internal knowledge stability in mitigating hallucinations remains underexplored. To address this gap, we propose SAKED, which mines stability-aware knowledge within LVLMs to dynamically enhance decoding and reduce hallucinations.
\vspace{-3mm}
\section{Methodology}
In this section, we study how instability in internal model knowledge contributes to hallucinations in LVLMs. We first formalize the LVLM generation process and introduce the necessary notations. We then analyze the relationship between knowledge stability and hallucination from three perspectives: attention heads, model layers, and decoding steps. Our empirical studies identify three key factors contributing to hallucinations: visual activation drift across attention heads, knowledge fluctuations across layers, and visual focus distraction between adjacent steps. {B}ased on these observations, we propose \textbf{S}tability-\textbf{A}ware \textbf{K}nowledge \textbf{E}nhanced \textbf{D}ecoding (\textbf{SAKED}). To quantify stability-aware knowledge, we design three metrics: Cross-Head Stability Score ({CHSS}), Cross-Layer Stability Score ({CLSS}), and Cross-Token Stability Score ({CTSS}). Finally, we identify stability-aware (positive) and stability-agnostic (negative) layer pair and incorporate them into a contrastive decoding strategy that suppresses decoding noise and improves faithfulness. SAKED dynamically corrects generated tokens and can be flexibly adapted to different models in a plug-and-play manner.
\subsection{Preliminaries} 
LVLMs typically consist of three key components: a vision encoder, a projection module, and an autoregressive language model. The vision encoder first converts an input image into a sequence of visual tokens {$\bm{\mathrm{x}}^{I} = \{\bm{x}^{I}_{1}, \bm{x}^{I}_{2}, \dots, \bm{x}^{I}_{m}\}$}.  In parallel, a text prompt is tokenized into \( n \) textual tokens {\( \bm{\mathrm{x}}^{P} = \{\bm{x}^{P}_{1}, \bm{x}^{P}_{2}, \dots, \bm{x}^{P}_{n}\} \)}. Here \(m\) and \(n\) are the lengths of the visual and textual tokens. The visual and textual embeddings are concatenated to form the model input \( \bm{\mathrm{x}} \), which is passed through an autoregressive language model composed of \( L \) stacked transformer decoder layers. At each layer \( l \), the model produces hidden states {\( \bm{\mathrm{h}}^l = \{ \bm{h}^l_0, \bm{h}^l_1, \dots, \bm{h}^l_{m+n-1} \} \)}. During generation, the hidden state at the final position \( \bm{h}^L_{m+n-1} \) is projected via an affine transformation \( \phi(\cdot) \) to produce a probability distribution over the vocabulary $\mathcal{V}$ ($\theta$ represents the LVLM parameters): 
\begin{equation}
    {\bm{y}_{t}\sim p_{\theta}(\bm{y}_{t}|\bm{\mathrm{x}}^{I}, \bm{\mathrm{x}}^{P}, \bm{y}_{<t})}
\end{equation}

\subsection{Visual Activation Drift across Attention Heads Induces Hallucinations}
We first examine the relationship between cross-head knowledge consistency and hallucination. In transformers, multi-head self-attention aggregates multiple “attention views” to form more robust representations. Figure~\ref{Attention} visualizes the visual activations of a grounded token (``\textcolor{OliveGreen}{bus}'') and a hallucinated token (``\textcolor{red!70!black}{people}'') across attention heads within a single layer. The grounded token ``\textcolor{OliveGreen}{bus}'' exhibits concentrated high-activation regions that align with its semantic meaning, whereas the hallucinated token ``\textcolor{red!70!black}{people}'' shows diffuse, higher-entropy activations. Moreover, the grounded token demonstrates stronger cross-head consistency in its activated regions than the hallucinated token. Together, these observations suggest that hallucinations are more likely when visual information varies substantially across attention heads. Motivated by these observations, we first define the Visual Activation Score (VAS) to quantify the activation strength and concentration level of the $k$-th attention head:
\begin{equation}
\label{activation}
\mathrm{VAS}_t^{(l,k)}= M_t^{(l,k)} + O_t^{(l,k)}
\end{equation}
At each generation step $t$, $M_t^{(l,k)}$ and $O_t^{(l,k)}$ denote the maximum value and entropy of visual attention map $A_t^{(l,k)}\in \mathbb{R}^{a\times a}$ in layer $l$, respectively. Heads with higher $\mathrm{VAS}$ are expected to be more informative and reliable.
As such, for each layer $l$, we select the $K$ most representative heads $\hat{A}_t^{(l,k)}$ with the highest $\mathrm{VAS}_t^{(l,k)}$ values {from $H$ heads, where $k\in\mathcal{K}$ and $\mathcal{K}=\{1,\ldots,K\}$. We further define the Cross-Head Stability Score ({CHSS}) to measure knowledge stability across attention heads:
\begin{equation}
\label{CHSS}
\mathrm{CHSS}_t^{l}= \frac{1}{K(K-1)}\sum_{s,k\in\mathcal{K},s< k}\mathrm{SoftIoU}(\hat{A}_t^{(l,s)}, \hat{A}_t^{(l,k)})
\end{equation}
We introduce $\mathrm{SoftIoU}$ to quantify the consistency between two selected heads $s$ and $k$ (We omit the subscript $t$ for brevity): 
\begin{equation}
\label{eq:softIoU}
\mathrm{SoftIoU}\!\left(\hat{A}^{(l,s)}, \hat{A}^{(l,k)}\right)
=
\frac{\left\|\min\!\left(\hat{A}^{(l,s)}, \hat{A}^{(l,k)}\right)\right\|_{1}}
{\left\|\max\!\left(\hat{A}^{(l,s)}, \hat{A}^{(l,k)}\right)\right\|_{1}+\varepsilon},
\end{equation}
Here, $\min(\cdot)$ and $\max(\cdot)$ are applied element-wise at the corresponding spatial locations in the attention map, and $\varepsilon$ is a small constant for numerical stability. 
\begin{figure}[ht]
  \begin{center}
  \centerline{\includegraphics[width=\columnwidth]{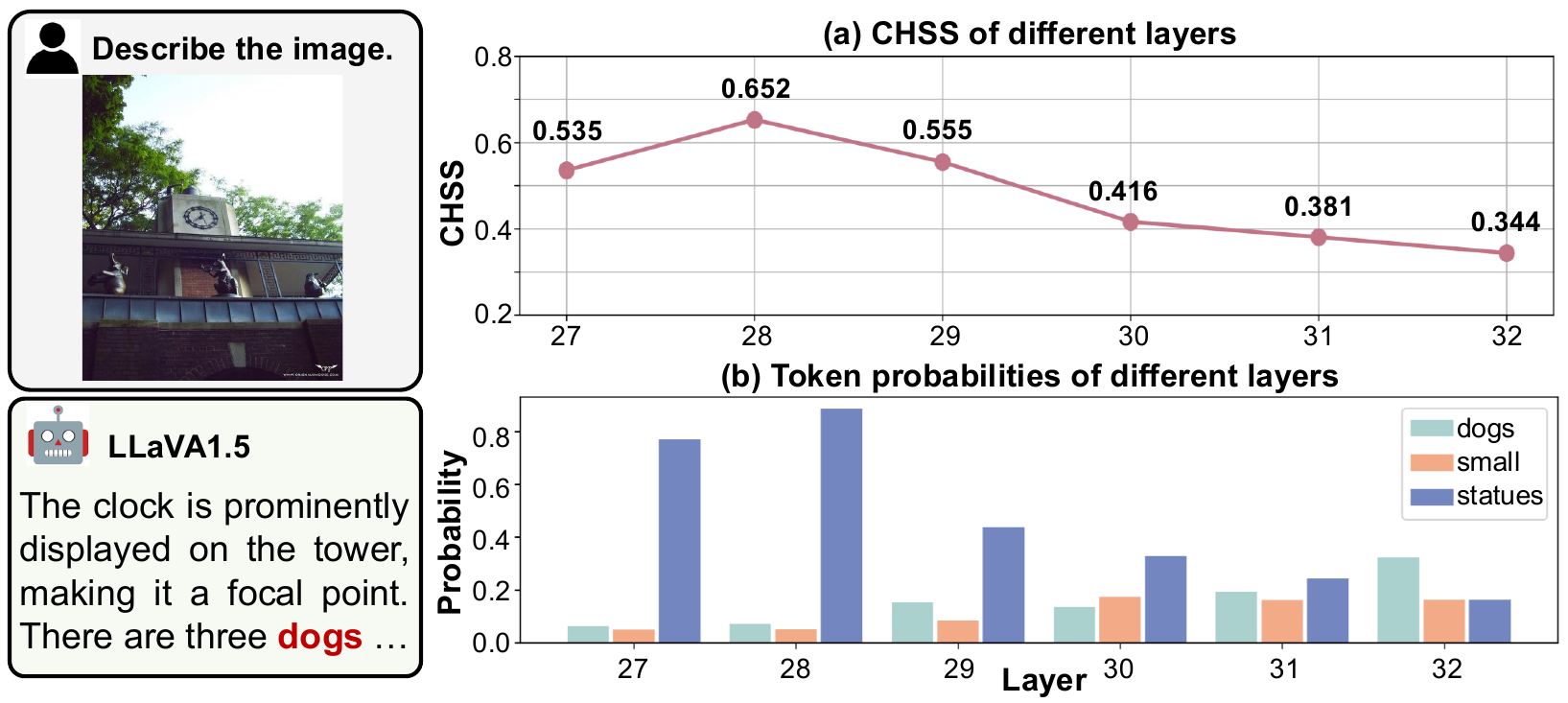}}
    \caption{(a) CHSS distribution across model layers. (b) Token probability distributions across model layers. For clarity, we show only the three highest probability tokens in the final layer: ``dogs'', ``small'', and ``statues''.
}
    \label{CHSSfig}
  \end{center}
  \vspace{-6mm}
\end{figure}

Figure~\ref{CHSSfig} presents the layer-wise distribution of $\mathrm{CHSS}_t^{l}$ together with the corresponding token probability distributions across layers. For clarity, we show only the three highest probability tokens (i.e., ``dogs'', ``small'', and ``statues'') in the final layer. We observe that higher $\mathrm{CHSS}$ values, such as those in layers 27 and 28, are associated with more accurate token predictions, indicating more consistent visual activation across heads. In contrast, the final layer exhibits the lowest $\mathrm{CHSS}$ and produces a hallucinated token. This observation motivates selecting the layer that contains the most stable knowledge to support more faithful token generation.


\subsection{Knowledge Fluctuations across Layers Implies Hallucinations}
In this subsection, we investigate how knowledge evolves and where knowledge instability happens across model layers. Figure~\ref{JSD_layer} visualizes the Jensen-Shannon Divergence (JSD) of layer-wise logit distributions. The left panel shows a representative example evaluated by LLaVA-1.5, where the hallucinated token ``\textcolor{red!70!black}{people}'' is highlighted. For each generated token $\bm{y}_t$, we apply LogitLens to the hidden state $\bm{h}_t^l$ from each layer and obtain the corresponding token distribution:
\begin{equation}
    \phi(\bm{h}_t^l) = \mathrm{softmax}(\mathrm{LogitLens}(\bm{h}_t^l)), ~~l\in L 
\end{equation}
\begin{figure}[ht]
  \begin{center}
  \centerline{\includegraphics[width=\columnwidth]{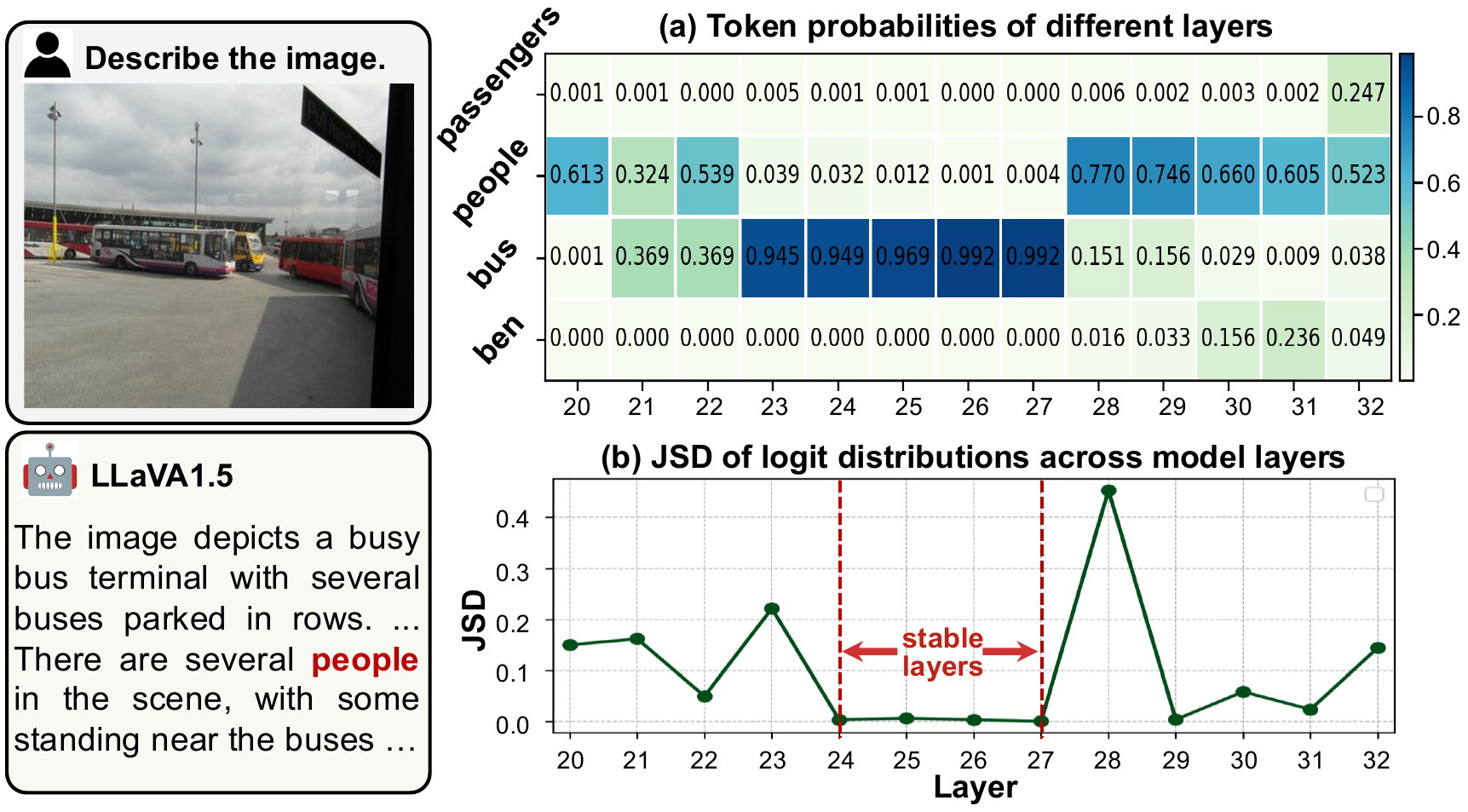}}
    \caption{
       (a) Probability distributions of the generated token ``\textcolor{red!70!black}{people}'' across model layers (due to space limitations, we show  the four highest-probability tokens in the final layer: ``passengers'', ``people'', ``bus'', and ``ben''); (b) JSD between the logit distributions of each layer and its adjacent layer, where lower JSD indicates a more stable flow of knowledge.
    }
    \label{JSD_layer}
  \end{center}
  \vspace{-6mm}
\end{figure}

\begin{figure*}[ht]
  \begin{center}
  \centerline{\includegraphics[width=\textwidth]{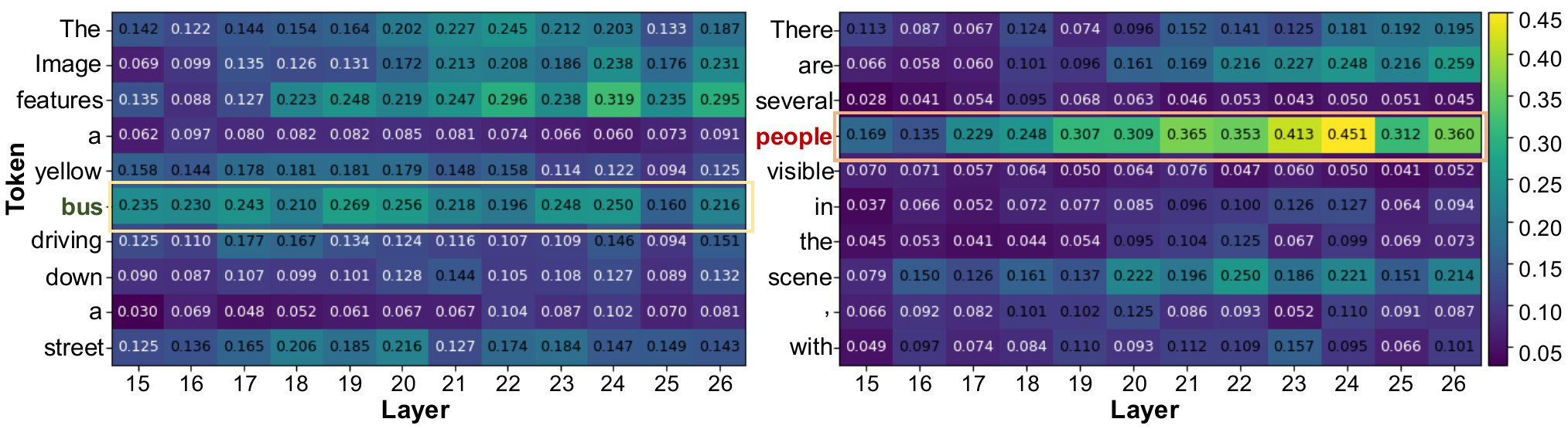}}
    \caption{
     VFD distributions of visual attention between each token and its adjacent token across model layers. The grounded token ``\textcolor{OliveGreen}{bus}'' and the hallucinated token ``\textcolor{red!70!black}{people}'' are highlighted in \textcolor{OliveGreen}{green} and \textcolor{red!70!black}{red}, respectively. Compared to real tokens, the hallucinated token exhibits substantial visual focus distraction during decoding.
    }
    \label{JSD_token}
  \end{center}
  \vspace{-6mm}
\end{figure*}
\vspace{-6mm}
LogitLens maps intermediate representations to the vocabulary space, enabling explicit visualization of how internal knowledge evolves across layers.
We then define the Cross-Layer Stability Score ({CLSS}) by measuring the JSD between consecutive layers:
\begin{equation}
\label{CLSS}
    \mathrm{CLSS}_t^l = 1-\mathrm{JSD}(\phi(\bm{h}^l_t),\phi(\bm{h}^{l-1}_t))
\end{equation}
Figure~\ref{JSD_layer} (a) illustrates probability distribution of the current token \bm{$y_t$} (``\textcolor{red!70!black}{people}'') changes across layers. Due to space limitations, we report only the four highest-probability tokens in the final layer (``passengers'', ``people'', ``bus'', and ``ben''). Figure~\ref{JSD_layer} (b) shows the JSD between adjacent layers, where lower values indicate more stable knowledge flow. Notably, the JSD increases sharply around layer 28, suggesting substantial internal knowledge fluctuation. In such cases, the model’s token preference becomes unstable across layers, which can lead to hallucinated outputs. We find that stable layer-to-layer transitions (layers 24-27) correlate with more faithful generations. Moreover, when hallucinations occur, the most stable transitions often appear in intermediate layers rather than in the final layer. 



\subsection{Visual Focus Distraction between Neighboring Tokens Triggers Hallucinations}
We further investigate how knowledge instability manifests across sequential decoding steps and contributes to hallucinations. Motivated by the observation that neighboring tokens often refer to the same visual entity. For example, in the phrase ``yellow bus'', both ``yellow'' and ``bus'' describe the entity bus. We thereby hypothesize that faithful decoding should maintain similar visual focus for adjacent tokens. Analogously, human decisions are more error-prone when attention is distracted. To test this hypothesis, we quantify Visual Focus Distraction (VFD) between consecutive generated tokens as the average JSD of their visual attention heads:
\begin{equation}
\mathrm{VFD}_t^{\,l}= \frac{1}{|\mathcal{H}|}\sum_{h\in|\mathcal{H}|}\mathrm{JSD}(A_t^{(l,h)}, A_{t-1}^{(l,h)})
\end{equation}
Figure~\ref{JSD_token} presents the VFD distributions of different tokens across layers. For the grounded token ``\textcolor{OliveGreen}{bus}'', the VFD remains consistently low, indicating that the model attends to consistent image regions when generating neighboring tokens. Its adjacent tokens ``yellow'' and ``driving'' exhibit similarly stable focus, consistent with their shared referent. In contrast, for the hallucinated token ``\textcolor{red!70!black}{people}'', the model’s visual focus deviates from that of the preceding token, resulting in substantially higher VFD values, particularly in deeper layers. This elevated divergence suggests increased uncertainty during decoding and indicates that visual focus distraction in deeper layers is associated with a higher risk of hallucination. Finally, we define the Cross-Token Stability Score ({CTSS}) to quantify model's temporal knowledge stability: 
\vspace{-2mm}
\begin{equation}
\label{CTSS}
\mathrm{CTSS}_t^{l}= 1-\mathrm{VFD}_t^{l}
\end{equation}
\begin{table*}[t]
\caption{Hallucination evaluation results on CHAIR across five models. Ci and Cs represent CHAIRi and CHAIRs, respectively. (OPERA cannot be applied to Qwen2.5VL and InternVL3 due to architectural constraints)}
\label{Chair}
\centering
\scalebox{0.90}{
\begin{tabular}{@{}c|cc|cc|cc|cc|cc|cc@{}}
\toprule
Model  & \multicolumn{2}{c|}{LLaVA1.5}
       & \multicolumn{2}{c|}{InstructBLIP}
       & \multicolumn{2}{c|}{MiniGPT-4}
       & \multicolumn{2}{c|}{InternVL3}
       & \multicolumn{2}{c|}{Qwen2.5VL}
       & \multicolumn{2}{c}{Average} \\ \midrule
Method & Ci $\downarrow$ & Cs $\downarrow$
       & Ci $\downarrow$ & Cs $\downarrow$
       & Ci $\downarrow$ & Cs $\downarrow$
       & Ci $\downarrow$ & Cs $\downarrow$
       & Ci $\downarrow$ & Cs $\downarrow$
       & Ci $\downarrow$ & Cs $\downarrow$ \\ \midrule
Greedy & 14.1 & 49.2 & \textbf{13.3} & 48.4 & 9.9 & 45.0 & 7.8 & 31.6 & 9.8 & 40.0 & 11.0 & 42.8 \\ 					
Beam   & 14.2 & 51.8 & 14.3 & 54.6 & 13.9 & 48.8 & 7.6 & 31.6 & 8.4 & 36.0 & 11.7 & 44.6 \\ 						
Dola   & 19.3 & 61.2 & 26.2 & 73.2 & 13.8 & 47.8 & 10.4 & 41.6 & 11.0 & 41.2 & 16.1 & 53.0 \\ 					
OPERA  & 12.8 & 44.6 & 14.2 & 46.4 & 12.8 & 44.6 & - & - & - & - & - & - \\ 									
Deco   & 12.5 & 46.4 & 13.5 & 46.4 & 9.4 & 32.2 & 8.0 & 33.2 & 9.4 & 38.8 & 10.6 & 39.4 \\  	
SAVER  & 12.8 & 40.4 & 13.6 & 47.8 & 10.0 & 32.8 & 8.3 & 32.8 & 8.4 & 35.6 & 10.5 & 37.4 \\ \midrule  
\textbf{SAKED}
       & \textbf{11.6} & \textbf{39.6}
       & 14.0 & \textbf{42.2}
       & \textbf{7.7} & \textbf{30.8}
       & \textbf{7.5} & \textbf{30.4}
       & \textbf{7.6} & \textbf{32.2}
       & \textbf{9.7} & \textbf{35.0} \\ \bottomrule
\end{tabular}%
}
\end{table*}

\begin{table*}[t]
\caption{Hallucination evaluation results on POPE across five models. The performance is the average F1 scores across the three evaluation settings: popular, random, and adversarial.}
\label{POPE}
\centering
\scalebox{0.95}{
\begin{tabular}{c|c|c|c|c|c|c}
\toprule
Method & LLaVA1.5 & InstructBLIP & MiniGPT-4 & InternVL3 & Qwen2.5VL & Average \\ \midrule
Greedy & 0.8220 & 0.8000 & 0.5850 & \textbf{0.9074} & 0.8136 & 0.7856 \\ 
Beam   & 0.8490 & 0.8440 & 0.7030 & 0.9021 & 0.8138 & 0.8224 \\  					
Dola   & 0.8320 & 0.8340 & 0.7280 & 0.8657 & 0.7489 & 0.8017 \\  					
Deco   & 0.8320 & 0.8492 & \textbf{0.7714} & 0.9069 & 0.8136 & 0.8346 \\  
SAVER  & 0.8489 & 0.8486 & 0.7680 & 0.9063 & 0.8137 & 0.8371 \\ \midrule
\textbf{SAKED} & \textbf{0.8536} & \textbf{0.8551} & 0.7639 & 0.9027 & \textbf{0.8721} & \textbf{0.8495} \\ \bottomrule
\end{tabular}%
}
\end{table*}


\begin{figure*}[ht]
  \begin{center}
  \centerline{\includegraphics[width=\textwidth]{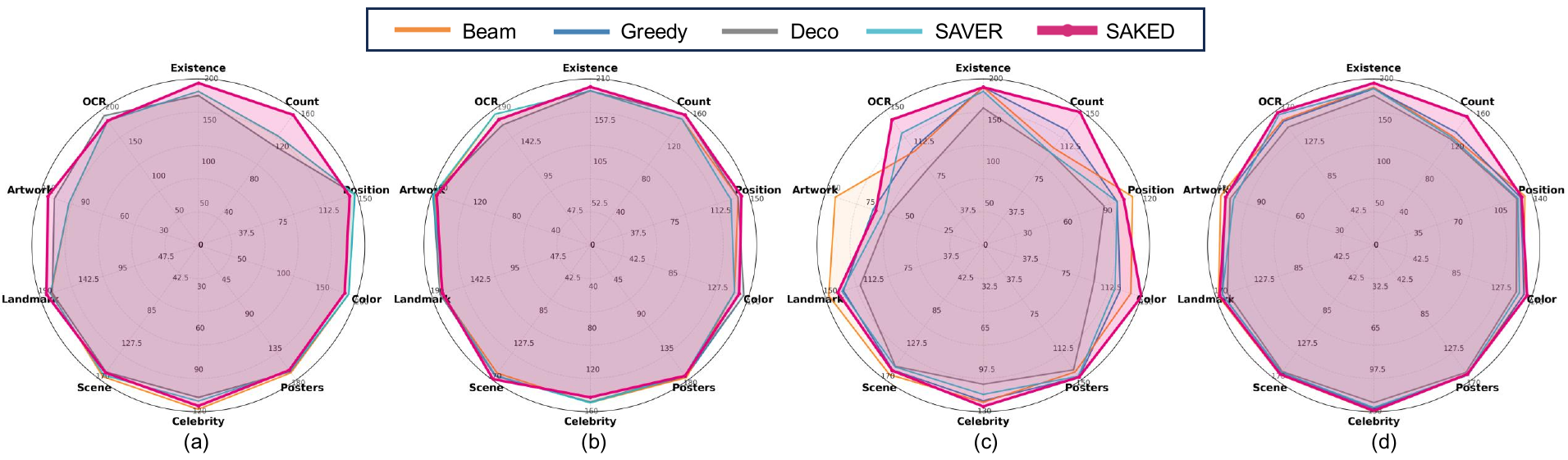}}
    \caption{
    Detailed MME evaluation results on 10 subsets: Existence, Count, Position, Color, Poster, Celebrity, Scene, Landmark, Artwork, and OCR on (a) Qwen2.5VL, (b) InternVL3, (c) LLaVA1.5, and (d) Average results on the three models.
    }
    \label{MME_FIGURE}
  \end{center}
  \vspace{-6mm}
\end{figure*}

\begin{table*}[t]
\caption{Hallucination evaluation results on AMBER. (Hal-Rate: hallucination rate; Cog: cognitive score; Avg. F1 indicates the average F1 scores across the following tasks: Discriminative, Existence, Attribute, State, Number, Action, and Relation.)}
\label{AMBER}
\centering
\scalebox{0.78}{
\begin{tabular}{@{}c|cccc|cccc|cccc@{}}
\toprule
Model  & \multicolumn{4}{c|}{Qwen2.5VL} & \multicolumn{4}{c|}{InternVL3} & \multicolumn{4}{c}{LLaVA1.5} \\ \midrule 
Method & CHAIR $\downarrow$ & Hal-Rate $\downarrow$ & Cog $\downarrow$ & Avg. F1 $\uparrow$
       & CHAIR $\downarrow$ & Hal-Rate $\downarrow$ & Cog $\downarrow$ & Avg. F1 $\uparrow$
       & CHAIR $\downarrow$ & Hal-Rate $\downarrow$ & Cog $\downarrow$ & Avg. F1 $\uparrow$ \\ \midrule
Greedy & 3.7 & 16.1 & 0.7 & 84.4 & 3.7 & 14.5 & \textbf{0.6} & 72.2 & 5.5 & 22.4 & 2.2 & 73.7 \\ 
Beam   & 3.6 & 15.6 & \textbf{0.6} & 84.9 & 3.2 & 12.7 & \textbf{0.6} & 73.3 & 5.5 & 21.0 & 2.1 & 74.5 \\ 
Dola   & 5.1 & 21.1 & 1.0 & 81.8 & 4.7 & 15.2 & 0.7 & 64.4 & 7.4 & 30.1 & 2.8 & \textbf{78.2} \\ 
Deco   & 3.7 & 16.1 & 0.7 & 84.4 & 3.6 & 14.5 & 0.7 & 72.5 & 5.8 & 21.8 & 2.3 & 60.8 \\ 
SAVER  & \textbf{3.5} & 15.6 & \textbf{0.6} & 84.7 & 4.1 & 16.1 & \textbf{0.6} & 71.5 & 6.1 & 22.5 & 2.1 & 68.7 \\ \midrule
\textbf{SAKED}
       & 3.8 & \textbf{14.4} & \textbf{0.6} & \textbf{86.5}
       & \textbf{2.7} & \textbf{11.0} & \textbf{0.6} & \textbf{74.1}
       & \textbf{4.9} & \textbf{20.0} & \textbf{1.9} & 75.4 \\ \bottomrule 	
\end{tabular}%
}
\end{table*}
\vspace{-6mm}

\begin{figure*}[ht]\begin{center}\centerline{\includegraphics[width=\textwidth]{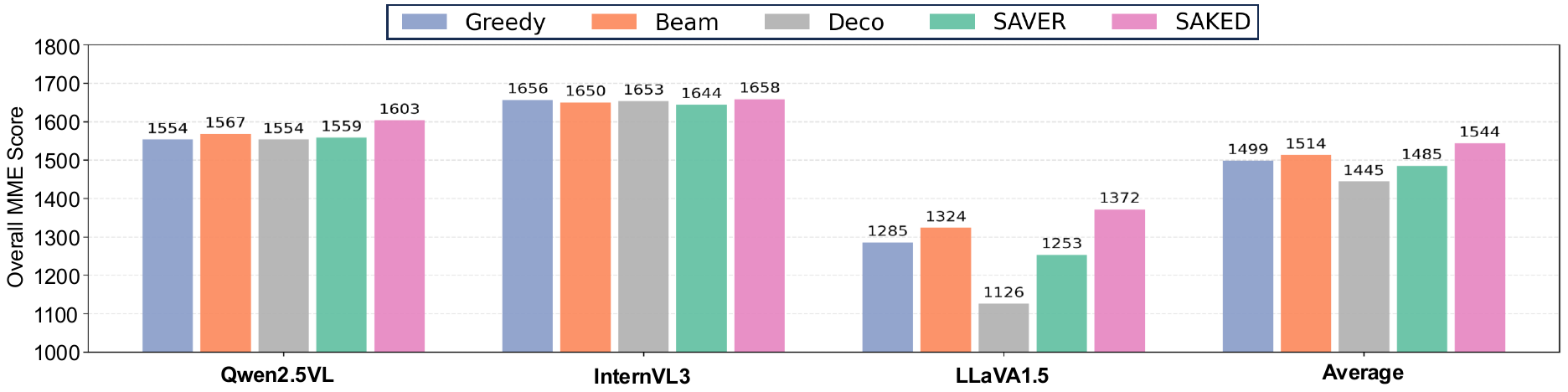}}
    \caption{Overall scores on Qwen2.5VL, InternVL3 and, LLaVA1.5 on MME full evaluation set. Higher scores indicate better general capability across perception and reasoning tasks.}
    \label{OverallMME_FIGURE}
  \end{center}
  \vspace{-6mm}
\end{figure*}

\begin{algorithm}[tb]
  \caption{Stability-Aware Knowledge Enhanced Decoding (SAKED)}
  \label{algo1}
  \begin{algorithmic}
    \STATE {\bfseries Require:} Inputs \{$\bm{\mathrm{x}^{I}}$, \bm{$\mathrm{x}^{P}$}\}, candidate layers ${L}_c$, contrastive weight $\alpha$, token revision weight $\beta$, score weights \{$\lambda_{1}$, $\lambda_{2}$, $\lambda_{3}$\} 
    \STATE At every decoding step $t$:
    \FOR{$l$ $\in$ ${L}_c$}
    \STATE Calculate $\mathrm{CHSS}_t^l$ using Eq.~\ref{CHSS} \textcolor{green!45!black}{\# Cross-Head Stability}
    \STATE Calculate $\mathrm{CLSS}_t^l$ using Eq.~\ref{CLSS} \textcolor{green!45!black}{\# Cross-Layer Stability}
    \STATE Calculate $\mathrm{CTSS}_t^l$ using Eq.~\ref{CTSS} \textcolor{green!45!black}{\# Cross-Token Stability}
    \STATE Overall \textbf{K}nowledge \textbf{S}tability \textbf{S}core ({KSS}):
    \STATE ${\mathrm{KSS}^{l}_t} = \lambda_{1}\mathrm{CHSS}^{l}_t + \lambda_{2}\mathrm{CLSS}^{l}_t + \lambda_{3}\mathrm{CTSS}^{l}_t$ 
    \ENDFOR
    \STATE Select positive layer $l^{+}$ and negative layer $l^{-}$: 
    \STATE ~~~~$l^{+}$ = $\underset{l\in{L}_c}{\mathrm{argmax}}$ ${\mathrm{KSS}^{l}_t}$, $l^{-}$ = $\underset{l\in{L}_c}{\mathrm{argmin}}$ ${\mathrm{KSS}^{l}_t}$
    \STATE Contrastive Decoding: 
    \STATE ~~~~$\bm{{y}_t^{cont}} = \mathrm{softmax}[(1+\alpha)\mathrm{logit}(\bm{h^{l^{+}}_t})-\alpha \mathrm{logit}(\bm{h^{l^{-}}_t)}]$
    \STATE Output Token Revision:
    \STATE ~~~~$\bm{\hat{y}_t}= \bm{{y}_t} + \beta\bm{{y}_t^{cont}}$ 
  \end{algorithmic}
\end{algorithm}
\subsection{Stability-Aware Knowledge Enhanced Decoding (SAKED)}
We define the overall Knowledge Stability Score (KSS) as a weighted sum of the three stability metrics introduced above, thereby quantifying knowledge stability from the head-, layer-, and token-level perspectives:
\begin{equation}
{\mathrm{KSS}^{l}_t} = \lambda_{1} \mathrm{CHSS}^{l}_t + \lambda_{2} \mathrm{CLSS}^{l}_t + \lambda_{3} \mathrm{CTSS}^{l}_t 
\end{equation}
The findings above indicate that internal knowledge stability varies across layers. Accordingly, we select the most stability-aware layer $l^{+}$ and the most stability-agnostic layer $l^{-}$ based on ${\mathrm{KSS}^{l}_t}$, forming a positive–negative layer pair:
\begin{equation}
\label{layer_selection}
l^{+} = \arg\max_{l \in {L}_c} {\mathrm{KSS}^{l}};~~~ l^{-} = \arg\min_{l \in {L}_c} {\mathrm{KSS}^{l}}
\end{equation}
We propose Stability-Aware Contrastive Decoding (SACD), which performs contrastive decoding by combining the logits computed from the hidden states $\bm{h}^{l^{+}}_t$ and $\bm{h}^{l^{-}}_t$ to obtain the next-token distribution:
\begin{equation}
\label{contrastive_decoding}
\bm{{y}}_t^{cont}= \mathrm{softmax}[(1+\alpha)\mathrm{logit}(\bm{h}^{l^{+}}_t)-\alpha \mathrm{logit}(\bm{h}^{l^{-}}_t)], 
\end{equation}
where $\alpha$ controls the contrast strength. Overall, stability-aware contrastive decoding suppresses decoding noise by dynamically leveraging the most reliable internal knowledge for faithful token generation. Finally, we dynamically adjust the output token by:
\begin{equation}
\label{revision}
\bm{\hat{y}}_t= \bm{{y}}_t + \beta\bm{{y}}_t^{cont},~~ \mathrm{subject}~\mathrm{to} ~ \bm{\hat{y}}_t\in\mathcal{V}_{candidate}
\end{equation}
Here, $\mathcal{V}$ is the whole vocabulary, and $\mathcal{V}_{candidate}$ indicates the top-$q$ (default: $q$=20) tokens with the highest probability under the original distribution $\bm{y}_t$, because we empirically observe that the grounded token remains within this set even when the model produces a hallucinated output. The pipeline of SAKED is summarized in Algorithm~\ref{algo1}. 

\section{Experiments}
\label{experiments}
\subsection{Implementation Details}
\label{implementation detail}

\textbf{Evaluation Models and Settings.}
We evaluate our proposed method and baseline approaches on five representative LVLMs: LLaVA-1.5~\cite{liu2023visual}, InstructBLIP~\cite{instructblip}, MiniGPT-4~\cite{zhu2023minigpt}, InternVL3\cite{zhu2025internvl3}, and Qwen2.5-VL\cite{bai2025qwen2}. Please refer to the Appendix~\ref{ExpSetup} for the detailed settings of our method.

\noindent \textbf{Baseline Methods.} We compare our method with two vanilla decoding strategies (greedy decoding and beam search) as well as four SOTA hallucination mitigation methods, detailed as follows:
Dola~\cite{chuang2023dola} is specifically designed for alleviating hallucinations in factual tasks for LLMs by reducing shallow semantic influences to improve the factuality of the final layer’s output. OPERA~\cite{huang2024opera} is a training-free decoding strategy that mitigates hallucinations in LVLMs by penalizing over-trusted summary tokens in self-attention during beam search and retrospectively reallocating token choices to better incorporate visual information. Deco~\cite{wang2024mllm} adaptively chooses relevant layers and integrates their knowledge into the final layer to adjust outputs. SAVER~\cite{li2025saver} leverages early-layer, token-level visual attention feedback to dynamically revise LVLM outputs, effectively reducing hallucinations. For all baselines, we either reproduce the methods using the latest model versions or follow the hyperparameter settings recommended in the officially released code to ensure fair comparisons.

\noindent \textbf{Benchmark and Metrics.}
We evaluate the effectiveness, generalizability, and language quality of our method across four challenging benchmarks:
$\bullet$ \textbf{CHAIR:} \cite{rohrbach2018object} Using the prompt  “\textit{Please describe the image in detail},” we assess instance- and sentence-level hallucination rates with CHAIR$\mathrm{i}$ and CHAIR$\mathrm{s}$. 
$\bullet$ \textbf{POPE:} \cite{li2023evaluating} Following the official POPE protocol, we report average F1 scores across three different settings as the primary metric. 
$\bullet$ \textbf{MME}~\cite{mme} is a practical benchmark encompassing a wide range of sub-tasks, including Existence, Count, Position, Color, Poster, Celebrity, Scene, Landmark, Artwork, and OCR, etc.
$\bullet$ \textbf{AMBER}~\cite{wang2023amber} evaluates hallucinations from more diverse perspectives such as attribute, relation, and existence. 

\subsection{Experiments on CHAIR}
Caption Hallucination Assessment with Image Relevance (CHAIR) \cite{rohrbach2018object} contains 500 images from the MSCOCO2014 validation set. It evaluates caption hallucination at both the instance-level and the sentence-level using the following metrics:  
\begin{equation}
\label{chair_i}
\mathrm{CHAIR_{i}} = \frac{|\{\mathrm{hallucinated~instances}\}|}{|\{\mathrm{all~mentioned~instances}\}|}
\end{equation}
\vspace{-2mm}
\begin{equation}
\label{chair_s}
\mathrm{CHAIR_{s}} = \frac{|\{\mathrm{hallucinated~captions}\}|}{|\{\mathrm{all~captions}\}|}
\end{equation}
Table \ref{Chair} reports hallucination performance on the CHAIR benchmark across five representative LVLMs, measured by CHAIRi (Ci) and CHAIRs (Cs), where lower values indicate fewer hallucinated objects and sentences. Greedy and Beam decoding exhibit consistently high hallucination rates across models, with average Ci/Cs of 11.0/42.8 and 11.7/44.6 respectively. 
Language-based method Dola ignores the visual information, thereby significantly degrading performance on various models. 
Hallucination-aware decoding methods (e.g., OPERA, Deco, SAVER) enhance models' visual perception and show partial improvements on some models. However, they fail to generalize consistently across architectures. And OPERA cannot be applied to all models due to architectural constraints. In contrast, our method SAKED achieves the lowest hallucination rates on almost all evaluated models, yielding the best average performance with Ci = 9.7 and Cs = 35.0, outperforming all baselines. These results validate that unstable internal knowledge commonly exists in different models, which induce hallucinations. By dynamically leveraging the most stable internal knowledge without additional training, SAKED achieves state-of-the-art hallucination mitigation performance and can be flexibly adapted to different architectures.

\vspace{-2.5mm}
\subsection{Experiments on POPE} 
\vspace{-1mm}
Unlike image captioning, the Polling based Object Probing Evaluation (POPE) \cite{li2023evaluating} is a VQA benchmark that contains 500 images from MSCOCO2014, with six questions paired with each image. POPE evaluates object hallucination by asking binary questions of the form ``Is there a \texttt{<object>} in the image?'' It includes three \texttt{<object>} settings: random (objects are sampled at random), popular (objects occur frequently), and adversarial (objects are closely related to those present in the image). Table~\ref{POPE} summarizes hallucination mitigation performance on POPE using the average F1 score across the random, popular, and adversarial settings. Prior methods such as Deco and SAVER improve visual perception, yet they remain inferior to our method. In contrast, SAKED yields consistent gains across backbones and achieves the highest average F1 score of 0.8495 among all compared methods. This consistency across diverse LVLM architectures suggests that hallucinations are not solely due to limited visual perception, but are also closely related to unstable knowledge aggregation.
\vspace{-2.5mm}
\subsection{Experiments on MME}
\vspace{-1mm}
MME \cite{mme} is a comprehensive evaluation benchmark designed to systematically assess the capabilities of LVLMs in understanding and reasoning over images. MME focuses on perception-oriented and cognition-oriented abilities with ``yes/no'' question format, covering a wide range of visual tasks such as Existence, Count, Position, Color, Poster, Celebrity, Scene, Landmark, Artwork, and OCR. Detailed results on specific tasks are shown in Figure~\ref{MME_FIGURE}, where our method consistently achieves outstanding evaluation scores on Qwen2.5VL, InternVL3 and LLaVA1.5 across different tasks. Figure~\ref{OverallMME_FIGURE} shows the overall scores across these tasks. SAKED consistently achieves the highest performance on all evaluated models in both perception-oriented and cognition-oriented tasks, yielding an average score of 1544, surpassing both the vanilla decoding strategy and previous hallucination mitigation methods. 

\begin{table}[t]
\caption{Ablation study of SAKED on CHAIR. Each row ablates one designed component of the proposed method.}
\label{Ablation}
\centering
\scalebox{0.67}{
\begin{tabular}{@{}c|cc|cc|cc|cc|cc@{}}
\toprule
Model & \multicolumn{2}{c|}{LLaVA1.5}
      & \multicolumn{2}{c|}{InstructBLIP}
      & \multicolumn{2}{c|}{InternVL3}
      & \multicolumn{2}{c|}{Qwen2.5VL}
      & \multicolumn{2}{c}{Average} \\ \midrule
Method & Ci $\downarrow$ & Cs $\downarrow$
       & Ci $\downarrow$ & Cs $\downarrow$
       & Ci $\downarrow$ & Cs $\downarrow$
       & Ci $\downarrow$ & Cs $\downarrow$
       & Ci $\downarrow$ & Cs $\downarrow$ \\ \midrule
w/o CHSS  & 13.5 & 44.8 & 14.1 & 42.4 & 8.0 & 33.0 & \textbf{7.0} & 32.4 & 10.7 & 38.1 \\ 							
w/o CLSS  & 12.1 & 41.0 & 17.5 & 48.4 & 8.3 & 32.4 & 7.5 & 32.4 & 11.4 & 38.6 \\ 							
w/o CTSS  & 12.4 & 42.4 & 15.1 & 47.6 & 7.8 & \textbf{30.2} & 8.0 & 34.2 & 10.8 & 38.6 \\ 						
w/o SACD  & 12.0 & 43.6 & 15.9 & 44.8 & 8.1 & 31.0 & 7.5 & 34.8 & 10.9 & 38.6 \\ \midrule	 						
\textbf{SAKED}
          & \textbf{11.0} & \textbf{38.0}
          & \textbf{14.0} & \textbf{42.2}
          & \textbf{7.5} & 30.4
          & 7.6 & \textbf{32.2}
          & \textbf{10.0} & \textbf{35.7} \\ \bottomrule
\end{tabular}%
}
\end{table}

\vspace{-2.5mm}
\subsection{Experiments on AMBER}
\vspace{-1mm}
Table \ref{AMBER} reports hallucination evaluation results on the AMBER benchmark across three LVLMs, measured by CHAIR, hallucination rate (Hal-Rate), cognitive score (Cog), and average F1 score over seven fine-grained tasks: Discriminative, Existence, Attribute, State, Number, Action, and Relation. Overall, SAKED demonstrates the most balanced and robust performance, achieving consistent reductions in hallucination-related metrics while maintaining outstanding task accuracy. By explicitly leveraging stability cues during decoding, SAKED effectively suppresses unreliable reasoning paths and achieves outstanding hallucination mitigation performance on both perception- and cognition-oriented metrics.
\vspace{-1mm}
\subsection{Ablation Studies}
\vspace{-1mm}
To evaluate the effectiveness of the proposed components, namely CHSS, CLSS, CTSS, and SACD, we report an ablation study of SAKED on the CHAIR benchmark across four representative LVLMs in Table~\ref{Ablation}. Removing any component consistently degrades hallucination mitigation performance, which validates the contribution of each module. The full SAKED framework achieves the best average results across the listed models (Ci=10.0 and Cs=35.7), indicating that cross-head, cross-layer, and cross-token stability modeling provides complementary benefits. In addition, removing SACD consistently worsens performance across all models, which supports its role in enhancing the decoding process. Overall, leveraging stability cues during token generation effectively helps reduce reliance on unstable internal knowledge and mitigates error propagation during decoding. 


We further analyze the impact of the score weights $\lambda_1$, $\lambda_2$, and $\lambda_3$. Since KSS is used only for layer selection, the relative contributions of CHSS, CLSS, and CTSS are critical. We fix $\lambda_1$ to 1.0 and tune $\lambda_2$ and $\lambda_3$ in Table~\ref{lambda}. The best performance is achieved with $\lambda_2 = 1.0$ and $\lambda_3 = 1.0$, yielding average Ci and Cs of 10.2 and 36.1, respectively. Additional ablations on the contrastive weight $\alpha$, the token revision weight $\beta$, and the candidate layer set ${L}_c$ are provided in Appendix~\ref{hyper}.    
\begin{table}[t]
\caption{Ablation study on score weights $\lambda_{1}$, $\lambda_{2}$, and $\lambda_{3}$, where $\lambda_{1}$ is fixed to 1.0.}
\label{lambda}
\centering
\scalebox{0.68}{
\begin{tabular}{@{}cc|cc|cc|cc|cc|cc@{}}
\toprule
 & & \multicolumn{2}{c|}{LLaVA1.5}
   & \multicolumn{2}{c|}{InstructBLIP}
   & \multicolumn{2}{c|}{InternVL3}
   & \multicolumn{2}{c|}{Qwen2.5VL}
   & \multicolumn{2}{c}{Average} \\ \midrule
$\lambda_{2}$ & $\lambda_{3}$
& Ci $\downarrow$ & Cs $\downarrow$
& Ci $\downarrow$ & Cs $\downarrow$
& Ci $\downarrow$ & Cs $\downarrow$
& Ci $\downarrow$ & Cs $\downarrow$
& Ci $\downarrow$ & Cs $\downarrow$ \\ \midrule

0.5 & 1.0  & 12.7 & 43.4 & 14.1 & 44.8 & 7.6& 30.2& 7.9& 33.2& 10.6& 37.9\\
1.0 & 0.5  & 12.0 & 41.6 & 17.7 & 45.6 & \textbf{7.4} & \textbf{28.0} & 7.7 & 32.8 & 11.2 & 37.0 \\
1.0 & 1.0  & \textbf{11.6} & \textbf{39.6} & \textbf{14.0} & \textbf{42.2} & 7.5 & 30.4 & \textbf{7.6} & \textbf{32.2} & \textbf{10.2} & \textbf{36.1} \\
1.0 & 2.0  & 13.0 & 42.0 & 14.9 & 45.0 & 7.6 & 30.4 & 8.2 & 36.4 & 10.9 & 38.5 \\
2.0 & 1.0  & 12.3 & 41.0 & 14.6 & 45.4 & \textbf{7.4}& 29.4& \textbf{7.6} & \textbf{32.2} & 10.5& 37.0\\

\bottomrule
\end{tabular}%
}
\end{table}

\vspace{-2mm}

\section{Conclusion}
\vspace{-2mm}
In this paper, we present {Stability-Aware Knowledge Enhanced Decoding} (SAKED), a training-free and plug-and-play decoding strategy for mitigating hallucinations in LVLMs. Through extensive analyses, we show that hallucinations are closely related to internal knowledge instability along three dimensions: visual information drift across attention heads, knowledge fluctuations across layers, and visual focus distraction between adjacent tokens. Building on these findings, we propose a Knowledge Stability Score (KSS) that quantifies layer-wise knowledge stability for each generated token by integrating CHSS, CLSS, and CTSS. We then use KSS to dynamically identify the most ``stable'' and most ``unstable'' layers and contrast their logits to further improve decoding faithfulness. Extensive experiments show that SAKED consistently enhances faithfulness across a wide range of tasks and benchmarks. SAKED requires neither external information nor model fine-tuning and can be readily adapted to different architectures. Overall, our study advances the understanding of hallucination mechanisms in LVLMs, and SAKED provides a practical step toward safer and more reliable multimodal generation.



\section*{Impact Statement}
This work tackles a key barrier to deploying LVLMs in real-world settings: hallucinated content that undermines trust and safety. We present a training-free, plug-and-play decoding strategy that mitigates hallucinations without requiring additional data, external knowledge sources, or any model fine-tuning. Beyond proposing a practical remedy, we provide a systematic analysis linking hallucinated outputs to internal knowledge instability. Guided by these insights, our method significantly improves generation faithfulness and strengthens the reliability of LVLM outputs, supporting safer and more trustworthy use across diverse applications. More broadly, our findings offer a principled lens for diagnosing hallucinations and may catalyze follow-up research toward more general, widely deployable approaches for improving LVLM reliability.

\bibliography{example_paper}

@String(AAAI = {AAAI})

@article{wang2023amber,
  title={Amber: An llm-free multi-dimensional benchmark for mllms hallucination evaluation},
  author={Wang, Junyang and Wang, Yuhang and Xu, Guohai and Zhang, Jing and Gu, Yukai and Jia, Haitao and Wang, Jiaqi and Xu, Haiyang and Yan, Ming and Zhang, Ji and others},
  journal={arXiv preprint arXiv:2311.07397},
  year={2023}
}

@article{jing2024fgaif,
  title={Fgaif: Aligning large vision-language models with fine-grained ai feedback},
  author={Jing, Liqiang and Du, Xinya},
  journal={arXiv preprint arXiv:2404.05046},
  year={2024}
}

@article{chuang2023dola,
  title={Dola: Decoding by contrasting layers improves factuality in large language models},
  author={Chuang, Yung-Sung and Xie, Yujia and Luo, Hongyin and Kim, Yoon and Glass, James and He, Pengcheng},
  journal={arXiv preprint arXiv:2309.03883},
  year={2023}
}

@inproceedings{liu2024paying,
  title={Paying more attention to image: A training-free method for alleviating hallucination in lvlms},
  author={Liu, Shi and Zheng, Kecheng and Chen, Wei},
  booktitle={European Conference on Computer Vision},
  pages={125--140},
  year={2024},
  organization={Springer}
}

@article{zhao2024harmonizing,
  title={Harmonizing visual text comprehension and generation},
  author={Zhao, Zhen and Tang, Jingqun and Wu, Binghong and Lin, Chunhui and Wei, Shu and Liu, Hao and Tan, Xin and Zhang, Zhizhong and Huang, Can and Xie, Yuan},
  journal={arXiv preprint arXiv:2407.16364},
  year={2024}
}

@article{yin2024survey,
  title={A survey on multimodal large language models},
  author={Yin, Shukang and Fu, Chaoyou and Zhao, Sirui and Li, Ke and Sun, Xing and Xu, Tong and Chen, Enhong},
  journal={National Science Review},
  volume={11},
  number={12},
  year={2024}
}

@article{chen2023shikra,
  title={Shikra: Unleashing multimodal llm's referential dialogue magic},
  author={Chen, Keqin and Zhang, Zhao and Zeng, Weili and Zhang, Richong and Zhu, Feng and Zhao, Rui},
  journal={arXiv preprint arXiv:2306.15195},
  year={2023}
}

@article{wang2024mllm,
  title={Mllm can see? dynamic correction decoding for hallucination mitigation},
  author={Wang, Chenxi and Chen, Xiang and Zhang, Ningyu and Tian, Bozhong and Xu, Haoming and Deng, Shumin and Chen, Huajun},
  journal={arXiv preprint arXiv:2410.11779},
  year={2024}
}

@inproceedings{leng2024mitigating,
  title={Mitigating object hallucinations in large vision-language models through visual contrastive decoding},
  author={Leng, Sicong and Zhang, Hang and Chen, Guanzheng and Li, Xin and Lu, Shijian and Miao, Chunyan and Bing, Lidong},
  booktitle={Proceedings of the IEEE/CVF Conference on Computer Vision and Pattern Recognition},
  pages={13872--13882},
  year={2024}
}

@inproceedings{zhang2024reflective,
  title={Reflective instruction tuning: Mitigating hallucinations in large vision-language models},
  author={Zhang, Jinrui and Wang, Teng and Zhang, Haigang and Lu, Ping and Zheng, Feng},
  booktitle={European Conference on Computer Vision},
  pages={196--213},
  year={2024},
  organization={Springer}
}

@inproceedings{wang2024mitigating,
  title={Mitigating fine-grained hallucination by fine-tuning large vision-language models with caption rewrites},
  author={Wang, Lei and He, Jiabang and Li, Shenshen and Liu, Ning and Lim, Ee-Peng},
  booktitle={International Conference on Multimedia Modeling},
  pages={32--45},
  year={2024},
  organization={Springer}
}

@article{zhou2023analyzing,
  title={Analyzing and mitigating object hallucination in large vision-language models},
  author={Zhou, Yiyang and Cui, Chenhang and Yoon, Jaehong and Zhang, Linjun and Deng, Zhun and Finn, Chelsea and Bansal, Mohit and Yao, Huaxiu},
  journal={arXiv preprint arXiv:2310.00754},
  year={2023}
}

@inproceedings{huang2024opera,
  title={Opera: Alleviating hallucination in multi-modal large language models via over-trust penalty and retrospection-allocation},
  author={Huang, Qidong and Dong, Xiaoyi and Zhang, Pan and Wang, Bin and He, Conghui and Wang, Jiaqi and Lin, Dahua and Zhang, Weiming and Yu, Nenghai},
  booktitle={Proceedings of the IEEE/CVF Conference on Computer Vision and Pattern Recognition},
  pages={13418--13427},
  year={2024}
}

@article{zhu2023minigpt,
  title={Minigpt-4: Enhancing vision-language understanding with advanced large language models},
  author={Zhu, Deyao and Chen, Jun and Shen, Xiaoqian and Li, Xiang and Elhoseiny, Mohamed},
  journal={arXiv preprint arXiv:2304.10592},
  year={2023}
}

@article{rohrbach2018object,
  title={Object hallucination in image captioning},
  author={Rohrbach, Anna and Hendricks, Lisa Anne and Burns, Kaylee and Darrell, Trevor and Saenko, Kate},
  journal={arXiv preprint arXiv:1809.02156},
  year={2018}
}

@misc{openai2023gpt4,
    title         = {GPT-4 Technical Report},
    author        = {OpenAI},
    year          = {2023},
    howpublished  = {\url{https://openai.com/research/gpt-4}},
    note          = {Accessed: 2024-10-18}
}

@misc{instructblip,
      title={InstructBLIP: Towards General-purpose Vision-Language Models with Instruction Tuning}, 
      author={Wenliang Dai and Junnan Li and Dongxu Li and Anthony Meng Huat Tiong and Junqi Zhao and Weisheng Wang and Boyang Li and Pascale Fung and Steven Hoi},
      year={2023},
      eprint={2305.06500},
      archivePrefix={arXiv},
      primaryClass={cs.CV}
}

@misc{mme,
      title={MME: A Comprehensive Evaluation Benchmark for Multimodal Large Language Models}, 
      author={Chaoyou Fu and Peixian Chen and Yunhang Shen and Yulei Qin and Mengdan Zhang and Xu Lin and Jinrui Yang and Xiawu Zheng and Ke Li and Xing Sun and Yunsheng Wu and Rongrong Ji},
      year={2024},
      eprint={2306.13394},
      archivePrefix={arXiv},
      primaryClass={cs.CV},
      url={https://arxiv.org/abs/2306.13394}, 
}

@inproceedings{li2023blip2,
  title={Blip-2: Bootstrapping language-image pre-training with frozen image encoders and large language models},
  author={Li, Junnan and Li, Dongxu and Savarese, Silvio and Hoi, Steven},
  booktitle={International conference on machine learning},
  pages={19730--19742},
  year={2023},
  organization={PMLR}
}

@article{liu2023visual,
  title={Visual instruction tuning},
  author={Liu, Haotian and Li, Chunyuan and Wu, Qingyang and Lee, Yong Jae},
  journal={Advances in neural information processing systems},
  volume={36},
  pages={34892--34916},
  year={2023}
}

@article{alayrac2022flamingo,
  title={Flamingo: a visual language model for few-shot learning},
  author={Alayrac, Jean-Baptiste and Donahue, Jeff and Luc, Pauline and Miech, Antoine and Barr, Iain and Hasson, Yana and Lenc, Karel and Mensch, Arthur and Millican, Katherine and Reynolds, Malcolm and others},
  journal={Advances in neural information processing systems},
  volume={35},
  pages={23716--23736},
  year={2022}
}

@article{awadalla2023openflamingo,
  title={Openflamingo: An open-source framework for training large autoregressive vision-language models},
  author={Awadalla, Anas and Gao, Irena and Gardner, Josh and Hessel, Jack and Hanafy, Yusuf and Zhu, Wanrong and Marathe, Kalyani and Bitton, Yonatan and Gadre, Samir and Sagawa, Shiori and others},
  journal={arXiv preprint arXiv:2308.01390},
  year={2023}
}

@article{li2025saver,
  title={SAVER: Mitigating Hallucinations in Large Vision-Language Models via Style-Aware Visual Early Revision},
  author={Li, Zhaoxu and Kong, Chenqi and Yu, Yi and Wu, Qiangqiang and Jiang, Xinghao and Cheung, Ngai-Man and Wen, Bihan and Kot, Alex and Jiang, Xudong},
  journal={The Association for the Advancement of Artificial Intelligence (AAAI)},
  year={2026}
}

@article{kalai2025language,
  title={Why language models hallucinate},
  author={Kalai, Adam Tauman and Nachum, Ofir and Vempala, Santosh S and Zhang, Edwin},
  journal={arXiv preprint arXiv:2509.04664},
  year={2025}
}

@inproceedings{yang2025mitigating,
  title={Mitigating hallucinations in large vision-language models via dpo: On-policy data hold the key},
  author={Yang, Zhihe and Luo, Xufang and Han, Dongqi and Xu, Yunjian and Li, Dongsheng},
  booktitle={Proceedings of the Computer Vision and Pattern Recognition Conference},
  pages={10610--10620},
  year={2025}
}

@article{kang2025see,
  title={See what you are told: Visual attention sink in large multimodal models},
  author={Kang, Seil and Kim, Jinyeong and Kim, Junhyeok and Hwang, Seong Jae},
  journal={arXiv preprint arXiv:2503.03321},
  year={2025}
}

@article{gong2024damro,
  title={Damro: Dive into the attention mechanism of lvlm to reduce object hallucination},
  author={Gong, Xuan and Ming, Tianshi and Wang, Xinpeng and Wei, Zhihua},
  journal={arXiv preprint arXiv:2410.04514},
  year={2024}
}

@inproceedings{favero2024multi,
  title={Multi-modal hallucination control by visual information grounding},
  author={Favero, Alessandro and Zancato, Luca and Trager, Matthew and Choudhary, Siddharth and Perera, Pramuditha and Achille, Alessandro and Swaminathan, Ashwin and Soatto, Stefano},
  booktitle={Proceedings of the IEEE/CVF Conference on Computer Vision and Pattern Recognition},
  pages={14303--14312},
  year={2024}
}

@inproceedings{jiang2024hallucination,
  title={Hallucination augmented contrastive learning for multimodal large language model},
  author={Jiang, Chaoya and Xu, Haiyang and Dong, Mengfan and Chen, Jiaxing and Ye, Wei and Yan, Ming and Ye, Qinghao and Zhang, Ji and Huang, Fei and Zhang, Shikun},
  booktitle={Proceedings of the IEEE/CVF Conference on Computer Vision and Pattern Recognition},
  pages={27036--27046},
  year={2024}
}

@article{sarkar2024mitigating,
  title={Mitigating Object Hallucination in MLLMs via Data-augmented Phrase-level Alignment},
  author={Sarkar, Pritam and Ebrahimi, Sayna and Etemad, Ali and Beirami, Ahmad and Ar{\i}k, Sercan {\"O} and Pfister, Tomas},
  journal={International Conference on Learning Representations},
  year={2025}
}

@inproceedings{wang2025shift,
  title={SHIFT: Smoothing Hallucinations by Information Flow Tuning for Multimodal Large Language Models},
  author={Wang, Sudong and Zhang, Yunjian and Zhu, Yao and Liu, Enci and Li, Jianing and Liu, Yanwei and Ji, Xiangyang},
  booktitle={Proceedings of the IEEE/CVF International Conference on Computer Vision},
  pages={3639--3649},
  year={2025}
}

@inproceedings{heiman2025factchexcker,
  title={FactCheXcker: Mitigating Measurement Hallucinations in Chest X-ray Report Generation Models},
  author={Heiman, Alice and Zhang, Xiaoman and Chen, Emma and Kim, Sung Eun and Rajpurkar, Pranav},
  booktitle={Proceedings of the Computer Vision and Pattern Recognition Conference},
  pages={30787--30796},
  year={2025}
}

@misc{gemini3,
  title = {Introducing Gemini 3},
  author = {{Google DeepMind}},
  howpublished = {\url{https://blog.google/products/gemini/gemini-3/}},
  year = {2025},
  note = {Accessed: 2025-12}
}

@misc{openai_introducing_gpt5_2025,
  title        = {Introducing GPT-5},
  author       = {OpenAI},
  howpublished = {\url{https://openai.com/index/introducing-gpt-5/}},
  year         = {2025},
  note         = {Accessed: 2025-12-20}
}

@article{bai2025qwen2,
  title={Qwen2. 5-vl technical report},
  author={Bai, Shuai and Chen, Keqin and Liu, Xuejing and Wang, Jialin and Ge, Wenbin and Song, Sibo and Dang, Kai and Wang, Peng and Wang, Shijie and Tang, Jun and others},
  journal={arXiv preprint arXiv:2502.13923},
  year={2025}
}

@article{zhu2025internvl3,
  title={Internvl3: Exploring advanced training and test-time recipes for open-source multimodal models},
  author={Zhu, Jinguo and Wang, Weiyun and Chen, Zhe and Liu, Zhaoyang and Ye, Shenglong and Gu, Lixin and Tian, Hao and Duan, Yuchen and Su, Weijie and Shao, Jie and others},
  journal={arXiv preprint arXiv:2504.10479},
  year={2025}
}

@article{li2023evaluating,
  title={Evaluating object hallucination in large vision-language models},
  author={Li, Yifan and Du, Yifan and Zhou, Kun and Wang, Jinpeng and Zhao, Wayne Xin and Wen, Ji-Rong},
  journal={arXiv preprint arXiv:2305.10355},
  year={2023}
}

@article{park2025second,
  title={SECOND: Mitigating Perceptual Hallucination in Vision-Language Models via Selective and Contrastive Decoding},
  author={Park, Woohyeon and Kim, Woojin and Kim, Jaeik and Do, Jaeyoung},
  journal={arXiv preprint arXiv:2506.08391},
  year={2025}
}

@article{xu2024reducing,
  title={Reducing tool hallucination via reliability alignment},
  author={Xu, Hongshen and Zhu, Zichen and Pan, Lei and Wang, Zihan and Zhu, Su and Ma, Da and Cao, Ruisheng and Chen, Lu and Yu, Kai},
  journal={arXiv preprint arXiv:2412.04141},
  year={2024}
}

@article{li2025hidden,
  title={The hidden life of tokens: Reducing hallucination of large vision-language models via visual information steering},
  author={Li, Zhuowei and Shi, Haizhou and Gao, Yunhe and Liu, Di and Wang, Zhenting and Chen, Yuxiao and Liu, Ting and Zhao, Long and Wang, Hao and Metaxas, Dimitris N},
  journal={arXiv preprint arXiv:2502.03628},
  year={2025}
}

@article{zhao2024mitigating,
  title={Mitigating object hallucination in large vision-language models via image-grounded guidance},
  author={Zhao, Linxi and Deng, Yihe and Zhang, Weitong and Gu, Quanquan},
  journal={arXiv preprint arXiv:2402.08680},
  year={2024}
}

@article{park2025steer,
  title={Steer LLM Latents for Hallucination Detection},
  author={Park, Seongheon and Du, Xuefeng and Yeh, Min-Hsuan and Wang, Haobo and Li, Yixuan},
  journal={arXiv preprint arXiv:2503.01917},
  year={2025}
}

@article{zou2024look,
  title={Look twice before you answer: Memory-space visual retracing for hallucination mitigation in multimodal large language models},
  author={Zou, Xin and Wang, Yizhou and Yan, Yibo and Lyu, Yuanhuiyi and Zheng, Kening and Huang, Sirui and Chen, Junkai and Jiang, Peijie and Liu, Jia and Tang, Chang and others},
  journal={arXiv preprint arXiv:2410.03577},
  year={2024}
}

@article{zhang2023language,
  title={How language model hallucinations can snowball},
  author={Zhang, Muru and Press, Ofir and Merrill, William and Liu, Alisa and Smith, Noah A},
  journal={arXiv preprint arXiv:2305.13534},
  year={2023}
}

@article{lv2024coarse,
  title={Coarse-to-fine highlighting: Reducing knowledge hallucination in large language models},
  author={Lv, Qitan and Wang, Jie and Chen, Hanzhu and Li, Bin and Zhang, Yongdong and Wu, Feng},
  journal={arXiv preprint arXiv:2410.15116},
  year={2024}
}

@article{chen2024halc,
  title={Halc: Object hallucination reduction via adaptive focal-contrast decoding},
  author={Chen, Zhaorun and Zhao, Zhuokai and Luo, Hongyin and Yao, Huaxiu and Li, Bo and Zhou, Jiawei},
  journal={arXiv preprint arXiv:2403.00425},
  year={2024}
}

@article{chen2024context,
  title={In-context sharpness as alerts: An inner representation perspective for hallucination mitigation},
  author={Chen, Shiqi and Xiong, Miao and Liu, Junteng and Wu, Zhengxuan and Xiao, Teng and Gao, Siyang and He, Junxian},
  journal={arXiv preprint arXiv:2403.01548},
  year={2024}
}

@article{wu2024evaluating,
  title={Evaluating and analyzing relationship hallucinations in large vision-language models},
  author={Wu, Mingrui and Ji, Jiayi and Huang, Oucheng and Li, Jiale and Wu, Yuhang and Sun, Xiaoshuai and Ji, Rongrong},
  journal={arXiv preprint arXiv:2406.16449},
  year={2024}
}

@inproceedings{zhualleviating,
  title={Alleviating Hallucinations in Large Language Models through Multi-Model Contrastive Decoding and Dynamic Hallucination Detection},
  author={Zhu, Chenyu and Liu, YeFeng and Zhang, Hao and Wang, Aowen and Chen, Guanhua and Wang, Longyue and Luo, Weihua and Zhang, Kaifu and others},
  booktitle={The Thirty-ninth Annual Conference on Neural Information Processing Systems}
}

@inproceedings{hesystematic,
  title={Systematic Reward Gap Optimization for Mitigating VLM Hallucinations},
  author={He, Lehan and Chen, Zeren and Shi, Zhelun and Yu, Tianyu and Shao, Jing and Sheng, Lu},
  booktitle={The Thirty-ninth Annual Conference on Neural Information Processing Systems}
}

@article{fang2025grounding,
  title={Grounding Language with Vision: A Conditional Mutual Information Calibrated Decoding Strategy for Reducing Hallucinations in LVLMs},
  author={Fang, Hao and Zhou, Changle and Kong, Jiawei and Gao, Kuofeng and Chen, Bin and Liang, Tao and Ma, Guojun and Xia, Shu-Tao},
  journal={arXiv preprint arXiv:2505.19678},
  year={2025}
}

@article{wang2025image,
  title={Image Tokens Matter: Mitigating Hallucination in Discrete Tokenizer-based Large Vision-Language Models via Latent Editing},
  author={Wang, Weixing and Ding, Zifeng and Gu, Jindong and Cao, Rui and Meinel, Christoph and de Melo, Gerard and Yang, Haojin},
  journal={arXiv preprint arXiv:2505.21547},
  year={2025}
}

@article{tan2025wisdom,
  title={Wisdom is Knowing What not to Say: Hallucination-Free LLMs Unlearning via Attention Shifting},
  author={Tan, Chenchen and Qu, Youyang and Li, Xinghao and Zhang, Hui and Cui, Shujie and Chen, Cunjian and Gao, Longxiang},
  journal={arXiv preprint arXiv:2510.17210},
  year={2025}
}

@inproceedings{fang2025enhancing,
  title={Enhancing Vision-Language Model Reliability with Uncertainty-Guided Dropout Decoding},
  author={Fang, Yixiong and Yang, Ziran and Chen, Zhaorun and Zhao, Zhuokai and Zhou, Jiawei},
  booktitle={The Thirty-ninth Annual Conference on Neural Information Processing Systems},
  year={2025}
}

@article{xia2026provable,
  title={Provable Robustness in Multimodal Large Language Models via Feature Space Smoothing},
  author={Xia, Song and Ding, Meiwen and Kong, Chenqi and Yang, Wenhan and Jiang, Xudong},
  journal={arXiv preprint arXiv:2601.16200},
  year={2026}
}

@article{kong2025moe,
  title={Moe-ffd: Mixture of experts for generalized and parameter-efficient face forgery detection},
  author={Kong, Chenqi and Luo, Anwei and Bao, Peijun and Yu, Yi and Li, Haoliang and Zheng, Zengwei and Wang, Shiqi and Kot, Alex C},
  journal={IEEE Transactions on Dependable and Secure Computing},
  year={2025},
  publisher={IEEE}
}
\bibliographystyle{icml2026}

\newpage
\appendix
\onecolumn

\begin{table}[ht]
\caption{Results of different maximum generation lengths on LLaVA-1.5 and Qwen2.5-VL. Our method consistently achieves lower Ci and Cs than vanilla decoding across all token budgets, with larger improvement at longer generation lengths.} 
\centering
\label{Length}
\resizebox{0.80\textwidth}{!}{%
\begin{tabular}{@{}c|cccccc|cccccc@{}}
\toprule
Model       & \multicolumn{6}{c|}{LLaVA1.5}                                                              & \multicolumn{6}{c}{Qwen2.5VL}                                                            \\ \midrule
Max. tokens & \multicolumn{2}{c}{64}       & \multicolumn{2}{c}{256}     & \multicolumn{2}{c|}{512}      & \multicolumn{2}{c}{64}     & \multicolumn{2}{c}{256}      & \multicolumn{2}{c}{512}      \\ \midrule
Method      & Ci $\downarrow$          & Cs $\downarrow$            & Ci $\downarrow$            & Cs $\downarrow$          & Ci $\downarrow$             & Cs $\downarrow$            & Ci $\downarrow$           & Cs $\downarrow$          & Ci $\downarrow$           & Cs $\downarrow$            & Ci $\downarrow$           & Cs $\downarrow$            \\ \midrule
Vanilla      & 7.3          & 22.8          & 14.4          & 48.8        & 14.1          & 49.2          & 6.3          & 15.0          & 8.9          & 36.6          & 9.8          & 40.0          \\
+SAKED        & \textbf{6.9} & \textbf{19.8} & \textbf{12.8} & \textbf{43.0} & \textbf{11.6} & \textbf{39.6} & \textbf{4.6} & \textbf{10.0} & \textbf{7.5} & \textbf{30.4} & \textbf{7.6} & \textbf{32.2} \\ \bottomrule
\end{tabular}%
}
\end{table}
\section{Experiment Setup}
\label{ExpSetup}

\subsection{Additional Implementation Details}
All experiments are conducted using deterministic decoding without sampling. We fix the decoding parameters, including top-$k$, top-$p$, and temperature to 1 for all datasets.
The maximum generation length is set to 512 tokens for CHAIR, 64 tokens for AMBER, and 10 tokens for POPE and MME, respectively. The beam search variant in our experiment essentially adopts a beam width of 5 while maintaining the others. Unless otherwise specified, all evaluated models are based on 7B-scale language backbones.



\subsection{Hardware Setup}
\label{hardware}
We run our benchmark and decoding method comparison experiments using one NVIDIA GPU with 48 GB of memory.


\section{Additional Experimental Results}
\subsection{Results of Different Maximum Token Numbers}
Table~\ref{Length} reports the results of different maximum generation lengths on LLaVA-1.5 and Qwen2.5-VL. Results are shown for maximum token limits of 64, 256, and 512, evaluated using two metrics, Ci and Cs from the CHAIR benchmark. We compare vanilla greedy decoding with our proposed method SAKED. Across all token budgets and both models, our method consistently outperforms greedy decoding, yielding lower Ci and Cs scores. The performance gain enhances as the maximum token limit increases, indicating that our approach scales more favorably with longer generations and better controls error accumulation in extended outputs.

\subsection{Language Quality Evaluations}
Table~\ref{caption-q-all} reports a comprehensive comparison of language quality among hallucination mitigation methods on three representative multimodal large language models: LLaVA-1.5, InternVL3, and InstructBLIP. We evaluate generation quality using BLEU-1/2/3/4, METEOR, and ROUGE-L, which jointly measure lexical overlap, n-gram consistency, and semantic alignment with ground-truth responses. Best and second-best results are highlighted in bold and underlined, respectively. 

On LLaVA-1.5, SAKED consistently outperforms all baselines across every metric, achieving the highest BLEU scores at all n-gram levels as well as the best METEOR and ROUGE-L results. This indicates that SAKED not only improves surface-level fluency but also enhances semantic fidelity while mitigating hallucinations. Dola and Deco provide moderate improvements over SAVER, but remain consistently below SAKED.
For InternVL3, Deco achieves the strongest overall performance, obtaining the best scores across all evaluation metrics. While SAKED does not surpass Deco on this model, it remains competitive and outperforms Dola on all metrics.
On InstructBLIP, SAKED again demonstrates superior performance, yielding the best BLEU-1/2, METEOR, and ROUGE-L scores, while achieving comparable BLEU-3/4 results to Deco and SAVER. These results highlight SAKED’s robustness in preserving both lexical accuracy and semantic consistency.

When averaging results across all three models, SAKED achieves the highest overall BLEU-1, BLEU-2, METEOR, and ROUGE-L scores, and remains highly competitive on  BLEU-3 and BLEU-4. This consistent improvement across diverse architectures suggests that SAKED provides a favorable balance between hallucination mitigation and language quality, avoiding the degradation in fluency or informativeness observed in some prior methods. 
\begin{table}[ht]
\centering
\caption{Language quality comparison of hallucination mitigation methods on LLaVA-1.5, InternVL3, and InstructBLIP. Results are reported in BLEU-1/2/3/4, METEOR, and ROUGE-L.}
\label{caption-q-all}
\resizebox{0.85\textwidth}{!}{%
\begin{tabular}{@{}c|c|cccccc@{}}
\toprule
Model & Method & BLEU-1 $\uparrow$ & BLEU-2 $\uparrow$ & BLEU-3 $\uparrow$ & BLEU-4 $\uparrow$ & METEOR $\uparrow$ & ROUGEL $\uparrow$ \\ \midrule
\multirow{4}{*}{LLaVA-1.5} & Dola   & \underline{18.9}  & \underline{11.0}  & 6.3   & 3.9   & \underline{31.8}   & \underline{16.0}   \\
&Deco   & 17.0  & \underline{11.0} & \underline{6.6}   & \underline{4.1}   & 29.8   & 15.5   \\
&SAVER  & 16.3  & 10.2  & 6.0   & 3.7   & 29.0   & 14.8   \\
&SAKED   & \cellcolor[HTML]{E2F0D9}\textbf{19.2}  & \cellcolor[HTML]{E2F0D9}\textbf{12.0}  & \cellcolor[HTML]{E2F0D9}\textbf{7.1}   & \cellcolor[HTML]{E2F0D9}\textbf{4.4}   & \cellcolor[HTML]{E2F0D9}\textbf{32.5}   & \cellcolor[HTML]{E2F0D9}\textbf{17.1}   \\ \midrule
\multirow{4}{*}{InternVL3} &Dola   & 11.5  & 7.0   & 3.7   & 2.2   & 22.8   & 10.3   \\
&Deco   & \textbf{14.2}  & \textbf{9.1}   & \textbf{5.3}   & \textbf{3.3}   & \textbf{26.5}   & \textbf{12.7}   \\
&SAVER  & \underline{13.9}  & \underline{8.6}   & \underline{4.9}   & \underline{3.0}   & \underline{26.4}   & \underline{12.3}   \\
&SAKED   & \cellcolor[HTML]{E2F0D9}12.4  & \cellcolor[HTML]{E2F0D9}7.8   & \cellcolor[HTML]{E2F0D9}4.3   & \cellcolor[HTML]{E2F0D9}2.6   & \cellcolor[HTML]{E2F0D9}24.0   & \cellcolor[HTML]{E2F0D9}11.1   \\ \midrule
\multirow{4}{*}{InstructBLIP} & Dola   & 14.0  & 8.1   & 4.6   & 2.8   & 26.7   & 12.1   \\
& Deco   & \underline{15.9}  & \underline{10.6}  & \textbf{6.6}  & \textbf{4.2}   & \underline{29.4}   & \underline{14.7}   \\
& SAVER  & 15.0  & {10.3}  & 6.4   & \underline{4.1}   & 28.2   & 14.3   \\
& SAKED   & \cellcolor[HTML]{E2F0D9}\textbf{16.6}  & \cellcolor[HTML]{E2F0D9}\textbf{10.7}  & \cellcolor[HTML]{E2F0D9}\underline{6.5}   & \cellcolor[HTML]{E2F0D9}\underline{4.1}   & \cellcolor[HTML]{E2F0D9}\textbf{30.4}   & \cellcolor[HTML]{E2F0D9}\textbf{14.9}   \\ \midrule
\multirow{4}{*}{Average} & Dola  & 14.8 & 8.7 & 4.9 & 3.0 & 27.1 & 12.8 \\
& Deco  & \underline{15.7} & \textbf{10.2} & \textbf{6.2} & \textbf{3.9} & \underline{28.6} & \underline{14.3} \\
& SAVER & 15.1 & \underline{9.7} & 5.8 & 3.6 & 27.9 & 13.8 \\
& SAKED & \cellcolor[HTML]{E2F0D9}\textbf{16.1} 
        & \cellcolor[HTML]{E2F0D9}\textbf{10.2} 
        & \cellcolor[HTML]{E2F0D9}\underline{6.0} 
        & \cellcolor[HTML]{E2F0D9}\underline{3.7} 
        & \cellcolor[HTML]{E2F0D9}\textbf{29.0} 
        & \cellcolor[HTML]{E2F0D9}\textbf{14.4} \\ \bottomrule
\end{tabular}%
}
\end{table}

\subsection{Additional Ablation Studies}
\label{hyper}
We conduct extensive ablation studies on three key hyper-parameters, $\alpha$, $\beta$, and ${L}_c$, across five representative LVLMs evaluated on the CHAIR benchmark. Following previous works~\cite{chuang2023dola, huang2024opera, li2025saver}, we set varying parameter values for different models. 

\subsubsection{Effect of the Weight $\alpha$ in SCAD}
Table~\ref{ab-alpha} summarizes the ablation results for the SCAD weighting factor $\alpha$, where {highlighted} cells correspond to our selected hyper-parameters. For LLaVA-1.5, increasing $\alpha$ consistently decreases the Ci metric, with the lowest value observed at $\alpha=0.5$, while Cs achieves its minimum at $\alpha=0.4$, suggesting that slightly larger $\alpha$ values are more beneficial for reducing instance-level hallucinations. InstructBLIP exhibits a similar pattern, where Ci is minimized at $\alpha=0.4$ and Cs at $\alpha=0.5$, indicating a trade-off between the two metrics.
InternVL3 demonstrates relatively stable Ci performance across different $\alpha$ values, whereas Cs reaches its optimal value at $\alpha=0.1$, implying that smaller $\alpha$ values are more effective in mitigating sentence-level hallucinations for this model. For Qwen2.5-VL, Ci is minimized at $\alpha=0.4$, while Cs attains its lowest value at $\alpha=0.3$, reflecting balanced sensitivity to $\alpha$. MiniGPT-4 achieves its best Cs performance at $\alpha=0.5$.

Overall, these findings suggest that moderate values of $\alpha$ generally provide a favorable balance between Ci and Cs across diverse architectures, while also highlighting the importance of model-specific tuning to further improve consistency performance.
\subsubsection{Effect of Final Output Revision Weight $\beta$}
Following the CHAIR benchmark, we investigate the impact of the token revision weight $\beta$ on both Ci and Cs as shown in Table~\ref{ab-beta}.

Overall, the results indicate that $\beta$ plays an important role in balancing instance-level and sentence-level hallucination. For LLaVA1.5, both Ci and Cs are minimized at $\beta=0.8$, suggesting that a moderate $\beta$ is effective in reducing hallucinations at both levels. InstructBLIP exhibits lower Ci and Cs at smaller $\beta$ values, while larger $\beta$ leads to increased hallucination, indicating higher sensitivity to this parameter.
For InternVL3, the lowest Ci and Cs are achieved at $\beta=0.4$, and performance remains relatively stable as $\beta$ increases, demonstrating robustness to different $\beta$ settings. Qwen2.5VL achieves the lowest Ci at $\beta=1.0$, while Cs is minimized at $\beta=0.8$ and $\beta=1.2$, reflecting a trade-off between instance-level and sentence-level hallucination. MiniGPT-4 shows the best instance-level consistency at $\beta=1.0$, whereas sentence-level hallucination is minimized at $\beta=0.4$.

\begin{table}[ht]
\centering
\caption{Ci and Cs results under different $\alpha$.}
\label{ab-alpha}
\resizebox{0.7\textwidth}{!}{%
\begin{tabular}{@{}c|cc|cc|cc|cc|cc@{}}
\toprule
\multirow{2}{*}[-0.2em]{$\alpha$} & \multicolumn{2}{c|}{LLaVA1.5}     & \multicolumn{2}{c|}{InstructBLIP} & \multicolumn{2}{c|}{InternVL3}    & \multicolumn{2}{c|}{Qwen2.5VL}    & \multicolumn{2}{c}{MiniGPT-4}      \\ \cmidrule(l){2-11} 
                          & Ci $\downarrow$ & Cs $\downarrow$ & Ci $\downarrow$ & Cs $\downarrow$ & Ci $\downarrow$ & Cs $\downarrow$ & Ci $\downarrow$ & Cs $\downarrow$ & Ci $\downarrow$ & Cs $\downarrow$ \\ \midrule
0.1                       & 12.4            & 41.4            & 14.2            & 47.0            & \cellcolor[HTML]{ADD8E6}7.6             & \cellcolor[HTML]{ADD8E6}{30.2}   & 7.3             & 32.6            & {7.7}    & 30.8            \\
0.2                       & 12.5            & 42.8            & 14.2            & 44.6            & 7.6             & 30.4            & 7.8             & 36.8            & 11.6            & 35.2            \\
0.3                       & 12.5            & 41.8            & 14.2            & 45.6            & 8.1             & 32.4            & \cellcolor[HTML]{ADD8E6}7.6             & \cellcolor[HTML]{ADD8E6}{32.2}   & 9.4             & 33.4            \\
0.4                       & \cellcolor[HTML]{ADD8E6}11.6            & \cellcolor[HTML]{ADD8E6}{39.6}   & {13.5}   & 43.2            & 8.1             & 33.4            & {7.1}    & 32.4            & 9.6             & 31.0            \\
0.5                       & {11.2}   & 39.8            & \cellcolor[HTML]{ADD8E6}13.6            & \cellcolor[HTML]{ADD8E6}{42.8}   & {7.5}    & 30.4            & 7.6&                 35.6& \cellcolor[HTML]{ADD8E6}9.0             & \cellcolor[HTML]{ADD8E6}{30.2}   \\ \bottomrule
\end{tabular}%
}
\end{table}

\begin{table}[ht]
\caption{Ci and Cs results under different $\beta$.}
\label{ab-beta}
\centering
\resizebox{0.7\textwidth}{!}{%
\begin{tabular}{@{}c|cc|cc|cc|cc|cc@{}}
\toprule
\multirow{2}{*}[-0.2em]{$\beta$} & \multicolumn{2}{c|}{LLaVA1.5}     & \multicolumn{2}{c|}{InstructBLIP} & \multicolumn{2}{c|}{InternVL3}    & \multicolumn{2}{c|}{Qwen2.5VL}    & \multicolumn{2}{c}{MiniGPT-4}      \\ \cmidrule(l){2-11} 
                         & Ci $\downarrow$ & Cs $\downarrow$ & Ci $\downarrow$ & Cs $\downarrow$ & Ci $\downarrow$ & Cs $\downarrow$ & Ci $\downarrow$ & Cs $\downarrow$ & Ci $\downarrow$ & Cs $\downarrow$ \\ \midrule
0.4                      & 12.9            & 43.0            & \cellcolor[HTML]{ADD8E6}13.6            & \cellcolor[HTML]{ADD8E6}42.8            & \cellcolor[HTML]{ADD8E6}{7.5}    & \cellcolor[HTML]{ADD8E6}{30.4}   & 8.7             & 38.6            & \cellcolor[HTML]{ADD8E6}9.6             & \cellcolor[HTML]{ADD8E6}{31.0}   \\
0.6                      & 12.8            & 43.0            & {14.1}   & {45.8}   & 9.3             & 32.4            & 7.9             & 37.2            & 11.0            & 35.4            \\
0.8                      & \cellcolor[HTML]{ADD8E6}{11.6}   & \cellcolor[HTML]{ADD8E6}{39.6}   & 14.8            & 46.8            & 7.9             & 30.8            & \cellcolor[HTML]{ADD8E6}7.6             & \cellcolor[HTML]{ADD8E6}{32.2}   & 9.8             & 34.2            \\
1.0                      & 14.4            & 43.6            & 17.9            & 46.4            & 8.6             & 32.0            & {7.4}    & 32.6            & {8.8}    & 36.6            \\
1.2                      & 12.7            & 41.8            & 17.4            & 46.0            & 8.0             & 32.4            & 8.8             & {32.2}   & 9.5             & 37.8            \\ \bottomrule
\end{tabular}%
}
\end{table}

In summary, these results suggest that moderate $\beta$ values provide a favorable balance between instance-level and sentence-level hallucination across different models, while extreme settings may amplify hallucinations. This demonstrates that the proposed method is effective and robust under a wide range of $\beta$ values when evaluated with CHAIR.


\subsubsection{Effect of Different Candidate Layers ${L}_c$}
Due to the variance of model architectures, we select different candidate layers ${L}_c$ for different models. 
Table~\ref{ab-Lc} shows that the optimal candidate layer range ${L}_c$ is model dependent, indicating that the most reliable stability contrast does not occur at a fixed depth across LVLM backbones. For LLaVA1.5, performance improves as ${L}_c$ shifts deeper, with the best results at 26-30 ($\mathrm{Ci}$=11.6, $\mathrm{Cs}$=39.6), suggesting that stability cues most useful for suppressing hallucinations emerge in late layers. InstructBLIP achieves its best performance in a mid-late range, 22-26 (13.6, 42.8), while both earlier and later ranges degrade results. In contrast, InternVL and Qwen2.5VL favor earlier mid ranges, namely 18-22 (9.6, 31.0) and 16-20 (7.2, 29.6), and their $\mathrm{Cs}$ increases noticeably when ${L}_c$ moves deeper, which is consistent with the hypothesis that later layers can amplify unstable knowledge aggregation or language priors, weakening faithfulness. MiniGPT-4 exhibits a mild tradeoff, where 16-20 yields the lowest $\mathrm{Ci}$ but 18-22 yields the lowest $\mathrm{Cs}$, suggesting that instance-level and sentence-level hallucination sensitivity can peak at slightly different depths. 

Overall, these results highlight that constraining ${L}_c$ to an appropriate mid or late block is important for reliably selecting $l^{+}$ and $l^{-}$, and they motivate choosing ${L}_c$ in a backbone-aware manner rather than adopting a universal default.

\subsection{Visualization Results}

We present qualitative examples comparing our method with the baseline greedy decoding strategy and the hallucination mitigation approach Deco. Our method significantly mitigates hallucinations by dynamically refining the decoding process using layer-wise stable knowledge while maintaining high-level quality. Figures~\ref{llava_vis}, \ref{instructblip_vis}, \ref{qwen_vis}, and \ref{internvl_vis} show representative results on LLaVA-1.5, InstructBLIP, Qwen2.5VL, and InternVL3, respectively.

\begin{table}[ht]
\caption{Ci and Cs results with different ${L}_c$.}
\label{ab-Lc}
\centering
\resizebox{0.9\textwidth}{!}{%
\begin{tabular}{@{}c|cc|cc|cc|c|cc|cc@{}}
\toprule
\multirow{2}{*}[-0.2em]{${L}_c$} & \multicolumn{2}{c|}{LLaVA1.5}                                   & \multicolumn{2}{c|}{InstructBLIP}                               & \multicolumn{2}{c|}{InternVL}                                  & \multirow{2}{*}[-0.2em]{${L}_c$} & \multicolumn{2}{c|}{Qwen2.5VL}                             & \multicolumn{2}{c}{MiniGPT-4}                                  \\ \cmidrule(lr){2-7} \cmidrule(l){9-12} 
                         & Ci $\downarrow$                & Cs $\downarrow$                & Ci $\downarrow$                & Cs $\downarrow$                & Ci $\downarrow$               & Cs $\downarrow$                &                          & Ci $\downarrow$             & Cs $\downarrow$              & Ci $\downarrow$               & Cs $\downarrow$                \\ \midrule
18-22                    & 14.7                           & 45.4                           & 14.9                           & 48.6                           & \cellcolor[HTML]{ADD8E6}{9.6} & \cellcolor[HTML]{ADD8E6}{31.0} & 16-20                    & \cellcolor[HTML]{ADD8E6}7.2 & \cellcolor[HTML]{ADD8E6}29.6 & 7.3                           & 33.8                           \\
20-24                    & 17.6                           & 46.2                           & 15.2                           & 48.8                           & 10.1                          & 32.8                           & 18-22                    & 7.5                         & 30.4                         & \cellcolor[HTML]{ADD8E6}{7.6} & \cellcolor[HTML]{ADD8E6}{32.2} \\
22-26                    & 13.9                           & 42.8                           & \cellcolor[HTML]{ADD8E6}{13.6} & \cellcolor[HTML]{ADD8E6}{42.8} & 11.0                          & 31.4                           & 20-24                    & 7.6                         & 30.2                         & 8.1                           & 35.4                           \\
24-28                    & 12.8                           & 41.2                           & 13.7                           & 47.2                           & 9.3                           & 37.8                           & 22-26                    & 8.1                         & 32.4                         & 9.0                           & 38.6                           \\
26-30                    & \cellcolor[HTML]{ADD8E6}{11.6} & \cellcolor[HTML]{ADD8E6}{39.6} & 14.2                           & 46.6                           & 10.3                          & 35.1                           & 24-28                    & 7.5                         & 31.4                         & 9.6                           & 41.8                           \\ \bottomrule
\end{tabular}%
}
\end{table}
\begin{figure*}[ht]

\centering

  \begin{center}
  
  \centerline{\includegraphics[width=1.0\textwidth]{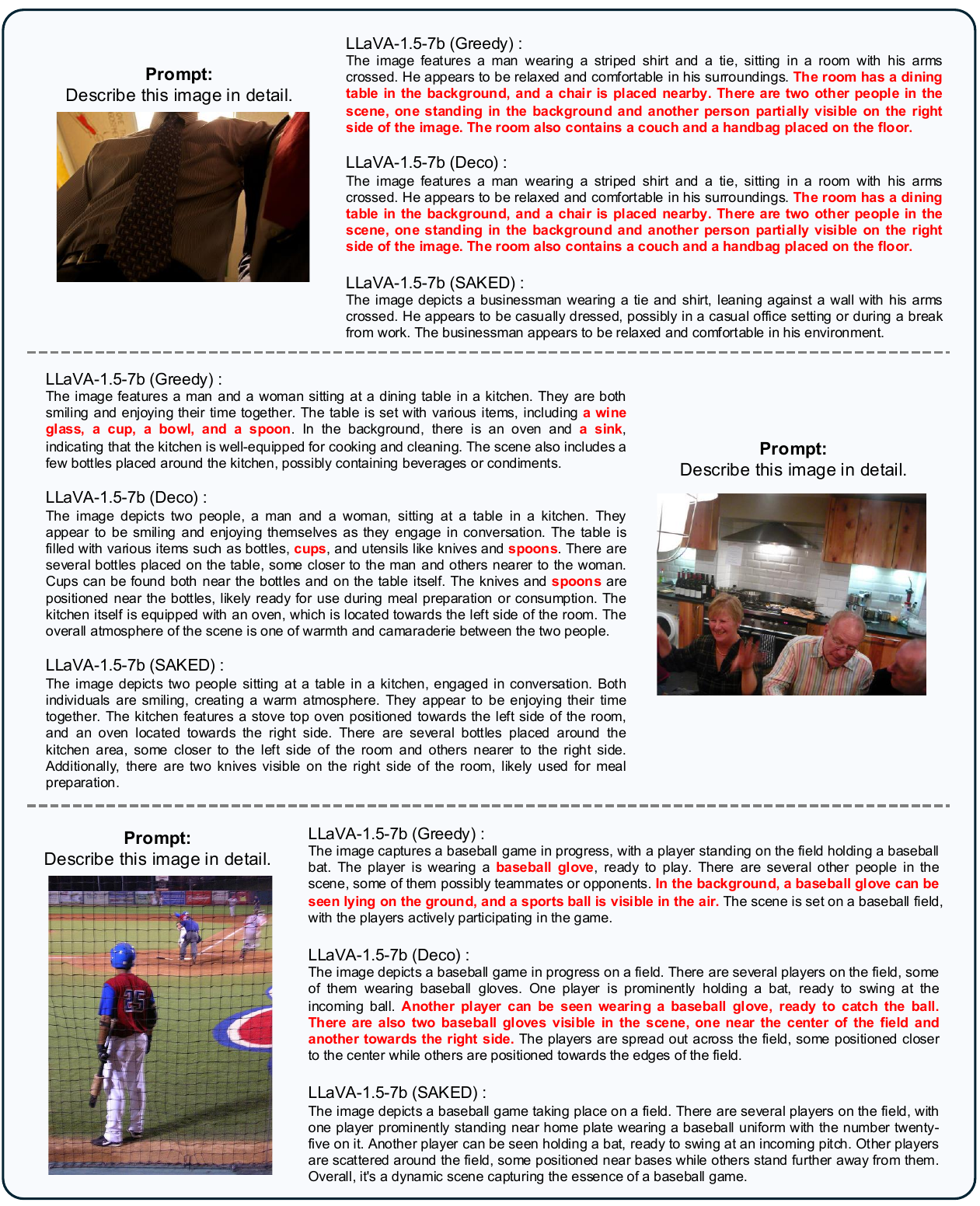}}
    \caption{ Qualitative visualization results on LLaVA-1.5 using different decoding methods, with hallucinated content highlighted in red.
    }
    \label{llava_vis}

  \end{center}
\end{figure*}
\begin{figure*}[ht]

\centering

  \begin{center}
  \centerline{\includegraphics[width=1.0\textwidth]{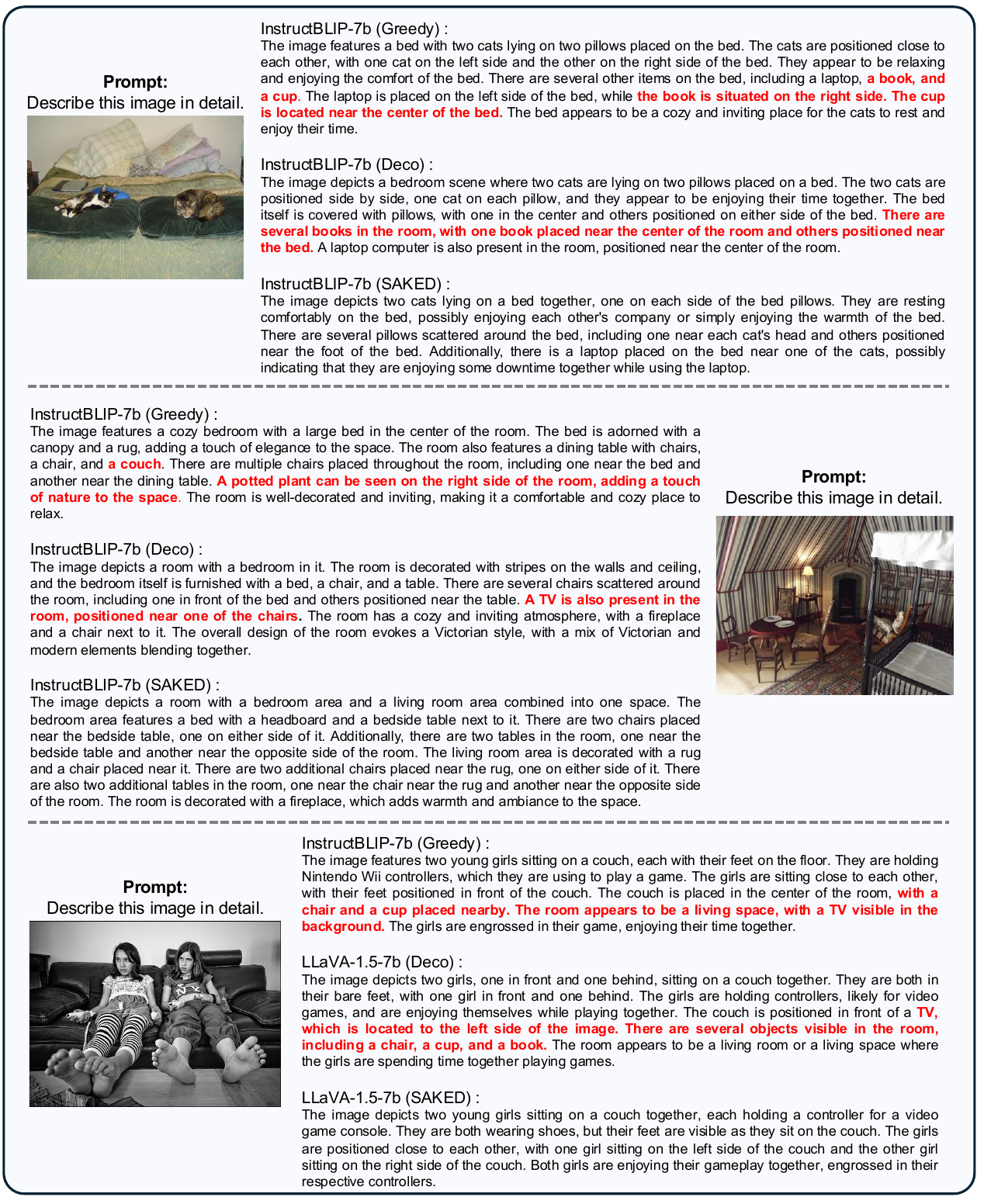}}
    \caption{ Qualitative visualization results on InstructBLIP using different decoding methods, with hallucinated content highlighted in red.
    }
\label{instructblip_vis}
  \end{center}
\end{figure*}
\begin{figure*}[ht]

\centering

  \begin{center}
  \centerline{\includegraphics[width=1.0\textwidth]{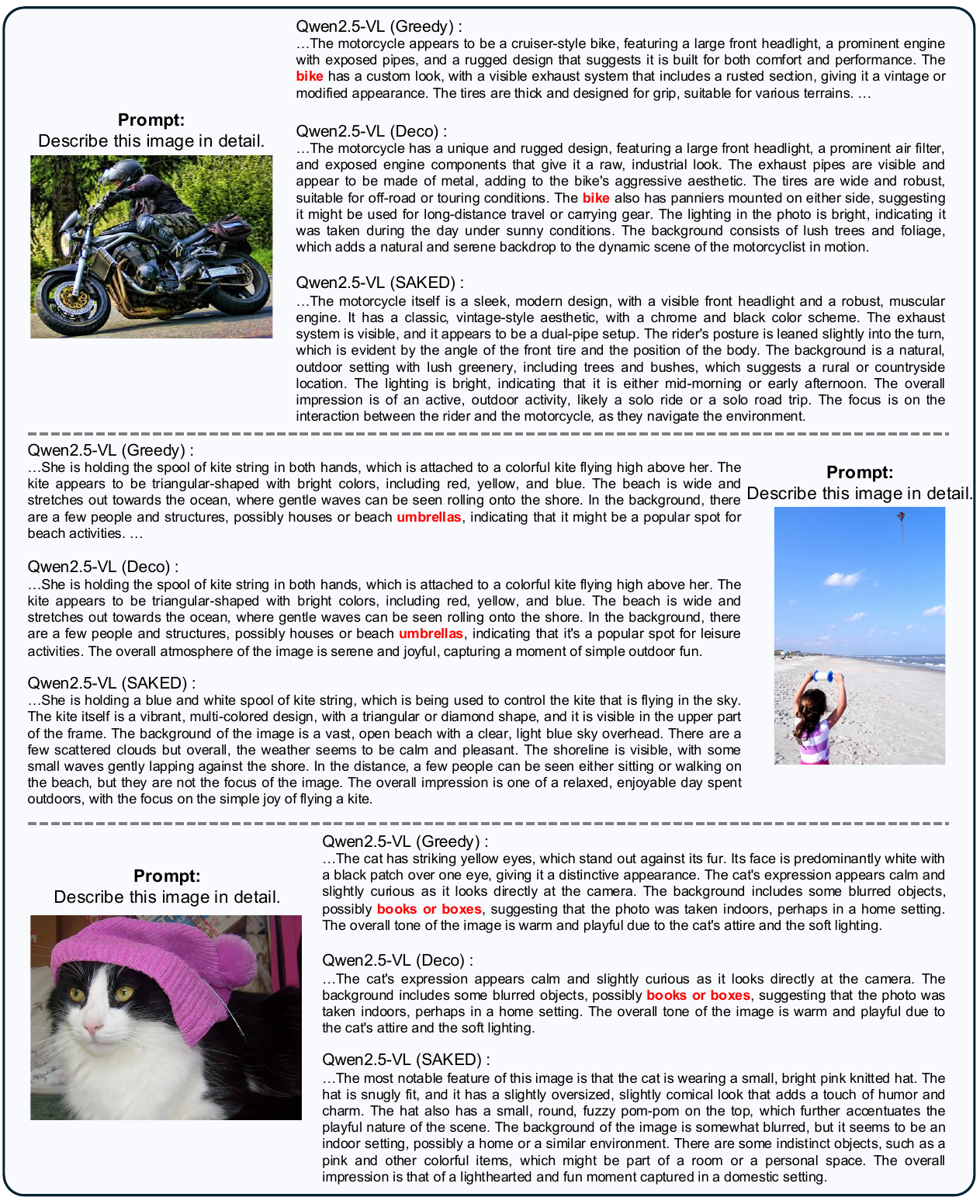}}
    \caption{ Qualitative visualization results on Qwen2.5VL using different decoding methods, with hallucinated content highlighted in red.
    }
\label{qwen_vis}
  \end{center}
\end{figure*}
\begin{figure*}[ht]

\centering

  \begin{center}
  \centerline{\includegraphics[width=1.0\textwidth]{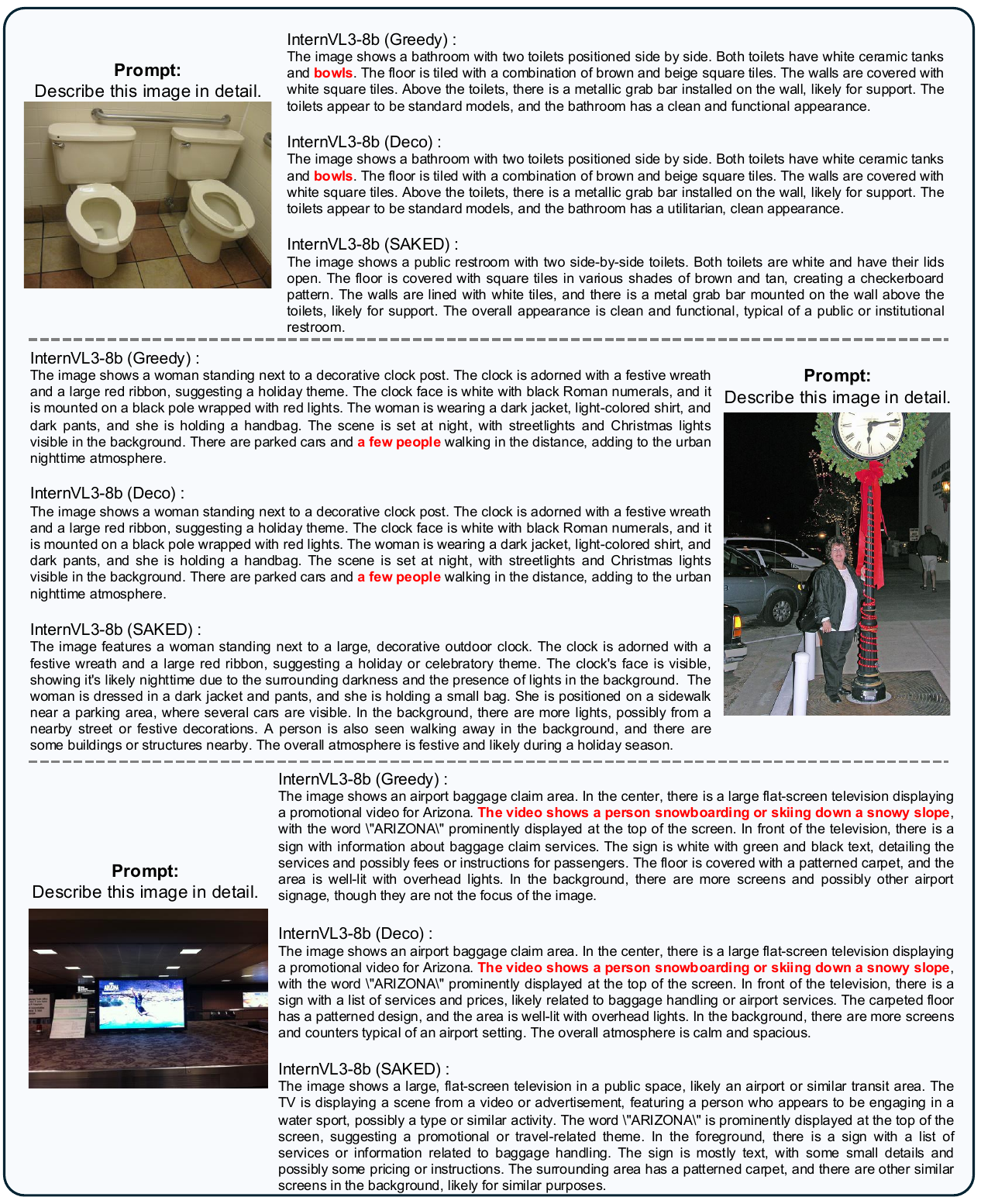}}
    \caption{ Qualitative visualization results on InternVL3 using different decoding methods, with hallucinated content highlighted in red.
    }
\label{internvl_vis}
  \end{center}
\end{figure*}

\section{Limitations and Future Work}
In the early stages of generation, token predictions are typically more stable, which reduces the extent to which temporal dependencies manifest in short sequences. As a result, CTSS has limited impact on short responses because it captures visual focus distraction between neighboring tokens, a signal that becomes more informative as temporal effects accumulate. Nevertheless, KSS also incorporates CHSS and CLSS, which exploit stable visual activation and layer-wise semantic cues to improve decoding even at early stages. As the sequence length increases, temporal dependencies accumulate, and SAKED remains effective at mitigating hallucinations in longer generations.

For future work, we will investigate more flexible ways to leverage CHSS, CLSS, and CTSS throughout the generation process. For example, these scores can be integrated into decoding with adaptive weighting that varies across generation stages, prediction confidence, or sequence length. Beyond inference time usage, CHSS, CLSS, and CTSS may also serve as auxiliary supervision signals during training, encouraging models to explicitly suppress behaviors that are prone to hallucination.

\end{document}